\documentclass{article}

\usepackage{microtype}
\usepackage{graphicx}
\usepackage{subcaption}
\usepackage{booktabs} % for professional tables

\usepackage{hyperref}

\usepackage[preprint]{icml2026}

\usepackage{amsmath}
\usepackage{amssymb}
\usepackage{mathtools}
\usepackage{amsthm}

\usepackage[capitalize,noabbrev]{cleveref}

\theoremstyle{plain}
\newtheorem{theorem}{Theorem}[section]

\newtheorem{lemma}[theorem]{Lemma}
\newtheorem{corollary}[theorem]{Corollary}
\theoremstyle{definition}

\newtheorem{assumption}[theorem]{Assumption}
\theoremstyle{remark}
\newtheorem{remark}[theorem]{Remark}

\usepackage{amsmath,amsfonts,bm}

\def\eqref#1{equation~\ref{#1}}
\def\1{\bm{1}}

\def\vu{{\bm{u}}}
\def\vv{{\bm{v}}}

\def\vx{{\bm{x}}}
\def\vy{{\bm{y}}}
\def\vz{{\bm{z}}}

\def\mA{{\bm{A}}}
\def\mB{{\bm{B}}}
\def\mC{{\bm{C}}}

\def\mI{{\bm{I}}}

\def\mM{{\bm{M}}}

\def\mP{{\bm{P}}}

\def\mS{{\bm{S}}}

\def\mU{{\bm{U}}}
\def\mV{{\bm{V}}}
\def\mW{{\bm{W}}}
\def\mX{{\bm{X}}}
\def\mY{{\bm{Y}}}

\def\mSigma{{\bm{\Sigma}}}

\DeclareMathAlphabet{\mathsfit}{\encodingdefault}{\sfdefault}{m}{sl}
\SetMathAlphabet{\mathsfit}{bold}{\encodingdefault}{\sfdefault}{bx}{n}

\def\gT{{\mathcal{T}}}

\def\gV{{\mathcal{V}}}

\def\sR{{\mathbb{R}}}

\newcommand{\E}{\mathbb{E}}

\newcommand{\R}{\mathbb{R}}

\DeclareMathOperator{\Tr}{Tr}

\usepackage{hyperref}
\usepackage{url}
\usepackage{enumitem}
\usepackage{graphicx}
\usepackage{subcaption}
\usepackage{caption}
\usepackage{float} 
\usepackage{listings}
\usepackage{wrapfig}
\usepackage{lipsum}

\def\vxi{{\bm{\xi}}}

\def\mTheta{{\bm{\Theta}}}
\def\mzero{{\bm{0}}}

\def\cN{{\mathcal{N}}}

\def\rank{{\textbf{rank}}}
\def\intdim{{\textbf{intdim}}}

\usepackage{amsmath}
\usepackage{amssymb}
\usepackage{mathtools}
\usepackage{amsthm}

\usepackage{bbm}
\usepackage{mathrsfs}

\usepackage[textsize=tiny]{todonotes}

\icmltitlerunning{LoRA is All You Need for Safety Alignment of Reasoning LLMs}

\begin{document}

\twocolumn[
  \icmltitle{LoRA is All You Need for Safety Alignment of Reasoning LLMs}

  % It is OKAY to include author information, even for blind submissions: the
  % style file will automatically remove it for you unless you've provided
  % the [accepted] option to the icml2026 package.

  % List of affiliations: The first argument should be a (short) identifier you
  % will use later to specify author affiliations Academic affiliations
  % should list Department, University, City, Region, Country Industry
  % affiliations should list Company, City, Region, Country

  % You can specify symbols, otherwise they are numbered in order. Ideally, you
  % should not use this facility. Affiliations will be numbered in order of
  % appearance and this is the preferred way.
  %\icmlsetsymbol{equal}{*}

  \begin{icmlauthorlist}
    \icmlauthor{Yihao Xue}{ucla}
    \icmlauthor{Baharan Mirzasoleiman}{ucla}
  \end{icmlauthorlist}

  \icmlaffiliation{ucla}{Department of Computer Science, University of California, Los Angeles}

  \icmlcorrespondingauthor{Yihao Xue}{yihaoxue@g.ucla.edu}

  % You may provide any keywords that you find helpful for describing your
  % paper; these are used to populate the "keywords" metadata in the PDF but
  % will not be shown in the document
  \icmlkeywords{Machine Learning, ICML}

  \vskip 0.3in
]

% this must go after the closing bracket ] following \twocolumn[ ...

% This command actually creates the footnote in the first column listing the
% affiliations and the copyright notice. The command takes one argument, which
% is text to display at the start of the footnote. The \icmlEqualContribution
% command is standard text for equal contribution. Remove it (just {}) if you
% do not need this facility.

% Use ONE of the following lines. DO NOT remove the command.
% If you have no special notice, KEEP empty braces:
\printAffiliationsAndNotice{}  % no special notice (required even if empty)
% Or, if applicable, use the standard equal contribution text:
% \printAffiliationsAndNotice{\icmlEqualContribution}

\begin{abstract}
Reasoning-capable LLMs have achieved major breakthroughs in solving complex problems, but recent work shows that acquiring and deploying strong reasoning can introduce significant safety risks. A common mitigation is to apply a secondary safety-alignment phase after reasoning is learned; however, safety alignment often degrades reasoning performance—a phenomenon known as the ``Safety Tax''. In this work, we show that a simple approach can largely bypass this trade-off: applying LoRA during SFT on refusal datasets. Despite its simplicity, this recipe achieves safety comparable to full-model alignment while preserving reasoning performance close to the original reasoning-tuned model, and the result holds across multiple model sizes and architectures, two safety benchmarks, and four reasoning benchmarks spanning mathematics, science, and code generation. We further ablate LoRA configurations and find that (1) rank-$1$ updates are sufficient to achieve the best safety–reasoning trade-off, (2) applying LoRA only to the MLP up-projection layers can outperform updating the full MLP, and (3) updating middle layers is more effective than updating early or late layers. Finally, we provide a theoretical analysis that helps understand when and why LoRA works, revealing that overshooting the rank budget (using a larger rank than needed for the finetuning task) induces base-task degradation at a rate inversely proportional to the intrinsic dimensionality of the base task. This suggests LoRA is most effective when the finetuning task is low-rank and the base capability is high-rank.\looseness=-1
\end{abstract}

\section{Introduction}

Large language models (LLMs) have made remarkable progress across a wide range of tasks. A major recent breakthrough is the emergence of LLMs with advanced reasoning capabilities, enabling them to solve complex problems previously out of reach. However, recent studies have reported significant safety risks associated with reasoning models \citep{jiang2025safechain,zhou2025hidden,huang2025safety,li2025smarter}.
Indeed, reasoning finetuning---the process through which LLMs acquire these capabilities---often compromises safety, even when starting from a safety-aligned checkpoint \citep{jiang2025safechain,zhou2025hidden,zhao2025trade,li2025smarter}.
For example, \citet{jiang2025safechain} show that models distilled for reasoning from DeepSeek-R1 become substantially less safe than their original base models.\looseness=-1

There has been significant effort in the literature to preserve LLMs' safety alignment during instruction finetuning. However, these approaches are not directly applicable to reasoning finetuning.
First, reasoning finetuning datasets are often highly curated \citep{muennighoff2025s1} and unlikely to contain unsafe content, limiting the applicability of data filtering methods \citep{shen2024seal,choi2024safety,bianchi2023safety}.
Second, methods that restrict model updates during finetuning \citep{hsu2024safe,mukhoti2023fine} are often ineffective in the reasoning setting, where acquiring reasoning capabilities typically requires longer training and more substantial weight updates than instruction finetuning.
To the best of our knowledge, the current literature does not offer a practical method for safety-preserving reasoning finetuning. Instead, the prevailing strategy is to apply a secondary safety alignment phase \emph{after} reasoning capabilities have been acquired. This phase---often implemented via supervised finetuning (SFT) or reinforcement learning (RL)---has become a standard step in modern LLM development. Although safety alignment finetuning can significantly improve model safety, it often comes at a steep cost to reasoning performance, a phenomenon referred to as the ``Safety Tax'' \citep{huang2025safety}. Even incorporating chain-of-thought (CoT) style reasoning into safety finetuning datasets \citep{jiang2025safechain} cannot fully preserve reasoning abilities \citep{huang2025safety}.

\begin{figure*}[!t]
    \centering
    \includegraphics[width=0.99\linewidth]{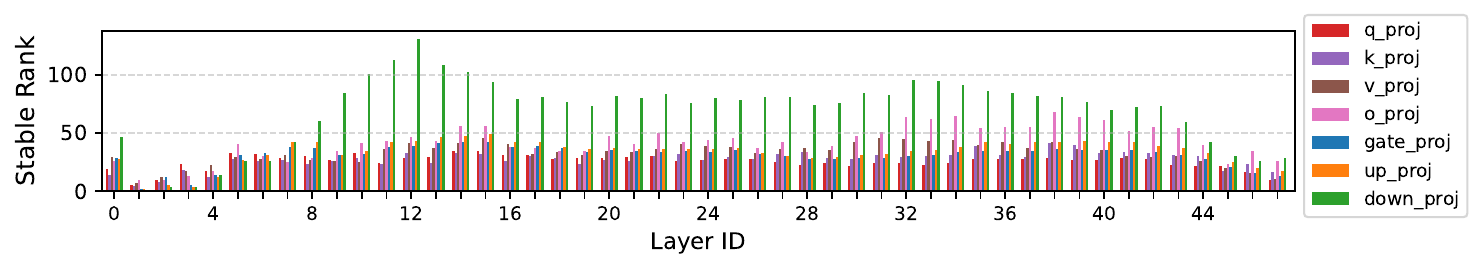}
     \vspace{-.12cm}
    \caption{We compute the stable rank of the difference between the full-model fine-tuned model’s weights and those of the original \texttt{DeepSeek-R1-Distill-Qwen-14B} for each layer. Here, the colors indicate the module types, and the x-axis shows the layer indices. We observe that the stable rank is quite high—ranging from around 40 to 100 for most layers}
    \label{fig:stable_rank}
    \vspace{-.2cm}
\end{figure*}

In this work, we investigate the algorithmic and structural factors underlying this trade-off. Existing evidence suggests that safety-related behavior in LLMs is often governed by a small number of dominant directions---either in activation space (e.g., steering vectors \citep{panickssery2023steering} and refusal features \citep{arditi2024refusal,yu2024robust}) or in weight space (e.g., \citet{jain2024makes,wei2024assessing} show that safety-critical weights tend to lie in a low-rank subspace). At the same time, we find that full-model safety alignment induces relatively high-rank changes across layers (Figure~\ref{fig:stable_rank}), which can unnecessarily perturb the reasoning solution and contribute to Safety Tax. This motivates a simple hypothesis: if safety can be achieved by modifying weights along a low-dimensional subspace, then restricting updates to that subspace should reduce interference with reasoning.\looseness=-1

\textbf{A simple recipe: LoRA for safety alignment of reasoning models.}
Guided by this hypothesis, we find the surprising effectiveness of a simple recipe for safety alignment of reasoning models: applying LoRA during SFT using a straightforward direct-refusal dataset. Despite its simplicity, this approach achieves safety performance on par with full-model alignment while preserving reasoning capability close to that of the original reasoning-tuned model. This result holds across three model sizes (7B, 8B, and 14B), two architectures (Qwen and Llama), two safety benchmarks covering a variety of harm categories, and four reasoning benchmarks spanning mathematics, science, and code generation. It represents a rare ``one stone, three birds'' outcome: strong safety, strong reasoning, and computational efficiency.\looseness=-1

\textbf{Understanding how much LoRA is sufficient.}
We further ablate the LoRA configuration to understand ``how much LoRA is sufficient''.
We make three findings.
(1) Setting the rank to 1 is sufficient (in terms of the Pareto frontier as the number of training epochs varies), so strong trade-off can be achieved at the lowest possible finetuning cost.
(2) Updating only the up-projection layers in the MLP yields an even better trade-off than updating the full MLP, while updating only the gate or down projections performs worse.
(3) Middle layers are more important for a good reasoning--safety trade-off: updating only 16 middle layers is often sufficient, whereas updating early or late layers yields worse results. Interestingly, this connects to prior findings that harmfulness-related representations often emerge in the middle layers of LLMs \citep{panickssery2023steering,arditi2024refusal}.
Overall, these results provide practical guidance on LoRA configurations that achieve strong performance at minimal cost.\looseness=-1

\textbf{Theory: when and why can LoRA avoid Safety Tax?}
Finally, we provide a theoretical explanation for when LoRA should be expected to succeed.
We analyze a stylized linear regression model with a base task (capturing the capability to preserve) and a finetuning task (capturing safety), and compare full-model finetuning to LoRA.
The theory captures the phenomenon that full-model finetuning fits the finetuning task but forgets the base task, whereas LoRA can perform well on both tasks.
Moreover, it yields two key insights about LoRA.
First, LoRA can fit the finetuning task well only when the dimensionality of the finetuning objective is comparable to the rank budget $r$.
Second, when $r$ exceeds the effective dimension required to represent the finetuning task, the resulting \emph{overshoot} (i.e., choosing $r$ larger than necessary) induces base-task degradation that scales roughly as\looseness=-1
\[
\begin{aligned}
&\text{base-task}\\[-0.7ex]
&\text{degradation}
\end{aligned}
\propto
\frac{\text{overshoot}}{\text{intrinsic dimension of the base capability}} .
\]
In other words, higher-dimensional base capabilities are substantially more tolerant to overshoot: increasing $r$ beyond what is required has a much smaller impact on base performance when the base capability is intrinsically high-dimensional.
Together, these results predict that LoRA is most effective when the finetuning objective is approximately low-dimensional while the base capability is intrinsically high-dimensional, making it tolerant to the inevitable mismatch in choosing $r$ in practice. We validate these predictions both through controlled numerical experiments and in a realistic setting. When safety-finetuning an instruction-tuned model (where the base capability is instruction following rather than reasoning), LoRA is no longer effective, consistent with the theory's requirement that the preserved base capability be intrinsically high-dimensional.

\section{Related Work}

To develop LLMs that are both safe and capable, models can be safety aligned before or after finetuning.

\textbf{Finetuning a safety-aligned model. } Finetuning a safety-aligned model
often leads to safety degradation. For instruction finetuning, safety degradation is shown across various model architectures and optimization strategies, including full-model and LoRA finetuning \citep{qi2023fine,hsiung2025llm,zhan2023removing}. To mitigate this issue, several methods have been proposed. \citet{shen2024seal,choi2024safety} focus on data %selection and 
filtering, aiming to remove unsafe examples from the finetuning data. \citet{bianchi2023safety} shows that injecting just a few hundred safety examples during instruction finetuning can improve safety. \citet{peng2025shape} leverages an existing guardrail model to encourage safe response segments while suppressing unsafe ones. \citet{lyu2024keeping} emphasizes the importance of prompt templates in preserving safety. Algorithmic approaches have also been explored: \citet{hsu2024safe} propose projecting LoRA updates into a ``safety subspace" derived from differences between aligned and unaligned models, while \citet{mukhoti2023fine} introduce regularization techniques to constrain changes in intermediate representations during finetuning. These approaches are, however, not applicable to reasoning finetuning. Acquiring reasoning capabilities typically requires longer training and more substantial weight updates compared to instruction finetuning. This results in losing the initial safety alignment of the model. Indeed, several recent studies have reported significant safety risks associated with reasoning-capable models \cite{jiang2025safechain,zhou2025hidden,huang2025safety,li2025smarter}.

\textbf{Safety alignment after finetuning.} 
Aligning a fine-tuned model is typically done via supervised finetuning (SFT) and/or reinforcement learning (RL) \citep{wei2021finetuned,griffith2013policy,dai2023safe,ouyang2022training,rafailov2023direct,bai2022training,guan2024deliberative}. For reasoning models, \citet{huang2025safety} identifies a key trade-off—termed the “safety tax”—where safety alignment can substantially degrade reasoning ability. \citet{jiang2025safechain} construct a safety finetuning dataset with long CoT responses, but the resulting models still show noticeable drops in reasoning performance \citep{huang2025safety}. Related work also studies how to mitigate over-refusal in safety-aligned models, such as post-hoc model editing \citep{dabas2025just} or inference-time methods \citep{ray2024mitigating,shi2024navigating,wang2024surgical}. However, these approaches do not directly address the trade-off induced by safety finetuning itself, and they primarily focus on instruction-following behavior rather than reasoning-specific capabilities. 
\looseness=-1

In our work, we show that a simple application of LoRA can effectively align reasoning models for safety without compromising their reasoning performance. We note that our work is specific to \textbf{safety alignment of reasoning models}, and should not be confused with either \textbf{utility finetuning of an already safety-aligned model} or \textbf{safety alignment of non-reasoning models}. The surprising effectiveness of LoRA appears to be tied to this specific setup and does not manifest as strongly in other settings, due to the interaction between LoRA and the tasks' inherent dimensionalities, as analyzed in our theory (Sec. \ref{sec:theory_linear}) and empirically observed in Fig. \ref{fig:non_reasoning}.\looseness=-1

\section{LoRA for Safety Alignment Without Compromising Reasoning}\label{sec: use_lora}

%Recent reasoning-capable LLMs have demonstrated remarkable power in solving highly complex tasks that were previously far beyond reach. However, a concerning trend shown in recent studies \cite{zhou2025hidden, jiang2025safechain, zhao2025trade} is that the acquisition of reasoning abilities is accompanied by a significant increase in risk: these models tend to comply with harmful prompts—such as those involving the design of biological weapons—posing serious threats to human society. To address this issue, 

We investigate whether the “safety tax” \citep{huang2025safety} can be mitigated—i.e., whether a model can be safety-aligned without sacrificing reasoning ability. We follow the setup of \citet{huang2025safety}, performing SFT on safety datasets that pair harmful requests with refusal responses.

Our key observation is that during full-model finetuning, which is used in \citep{huang2025safety}, the weights undergo relatively high-rank changes. As shown in Figure~\ref{fig:stable_rank}, we observe high stable ranks for the weight updates, i.e., differences between the fine-tuned model’s weights and those of the initial model, in most layers. However, prior evidence suggests that safety behavior in LLMs is typically governed by only a single or a few directions in the activations \citep{panickssery2023steering,arditi2024refusal} and weights \citep{wei2024assessing,jain2024makes}, indicating that a small low-rank modification may be sufficient to induce safe behavior, without altering the entire weight space. Thus, we conjecture that the degradation in reasoning performance is caused by full-model finetuning introducing unnecessary changes in many directions, which interfere with critical weights responsible for reasoning.

To address this, we consider Low-Rank Adaptation (LoRA) \citep{hu2022lora}, originally proposed as a parameter-efficient finetuning method to reduce training cost and memory usage. Rather than updating the full weight matrices, LoRA injects trainable low-rank matrices into existing layers while keeping the original weights frozen. Formally, a weight matrix $\mW \in \mathbb{R}^{d \times k}$ is modified as:\looseness=-1
\begin{equation}
\label{eq: lora}
 \mW' = \mW + \Delta \mW, ~~~~\text{where}~~~~ \Delta\mW= \frac{\alpha}{r}\mB\mA,   
\end{equation}
where $\mB \in \mathbb{R}^{d \times r}$ and $\mA \in \mathbb{R}^{r \times k}$ are the trainable low-rank matrices with $r\ll \min(d,k)$, and $\frac{\alpha}{r}$ is the scaling factor, with $\alpha$ being a hyperparameter.

% LoRA is particularly well-suited to our needs: it restricts updates to a low-rank subspace, thereby significantly reducing interference with the original weights. Additionally, LoRA allows updates to be applied to specific modules within the model, leaving the rest untouched. This modularity enables fine-grained control over which components to modify, further minimizing interference with reasoning-related weights. 

 LoRA is particularly well-suited to our needs: it restricts updates to a low-rank subspace, thereby significantly reducing interference with the original weights. We will show in our experiments that this method works excellently, enabling the model to become safe while maintaining strong reasoning performance across benchmarks. As an additional benefit, LoRA is significantly more computationally efficient than full-model finetuning.

%Based on this reasoning, we propose a simple and effective approach: using LoRA for safety alignment finetuning. As we will show in our experiments, this method works surprisingly well, enabling the model to become safe without compromising its reasoning capabilities. As an additional benefit, LoRA is significantly more computationally efficient than full-model finetuning.

\section{LoRA Bypasses the ``Safety Tax"}\label{sec:lora_bypass}

In this section, we first introduce our safety finetuning and evaluation pipeline. Then, we evaluate models' safety alignment and reasoning performance after full-model and LoRA safety finetuning.

\subsection{Safety Alignment Fine-Tuning of Reasoning LLMs}

We begin with a reasoning-capable language model and perform safety-alignment finetuning via SFT on a dataset of harmful questions paired with refusal responses, training the model to reject harmful requests. Following \cite{huang2025safety}, we choose SFT over RL-based methods because it is simpler, more cost-effective, and avoids the need for additional components such as a reward model.
\looseness=-1
% Although prior work \citep{huang2025safety} has shown that such SFT-based finetuning can significantly degrade a model’s reasoning ability, we will demonstrate that, interestingly, when using LoRA-based optimization, the model's reasoning capability can be retained.

In our training setup, we compare (1) \textbf{full-model finetuning}, as in \citep{huang2025safety}, where all model parameters are updated using standard gradient-based optimization; and (2) \textbf{LoRA finetuning} described in Section~\ref{sec: use_lora}.\looseness=-1

%LoRA (Low-Rank Adaptation) is a parameter-efficient finetuning technique originally proposed by \cite{hu2022lora} to reduce training cost and memory usage. Instead of updating the full weight matrices, LoRA injects trainable low-rank matrices into the existing layers of the model, keeping the original weights frozen. Formally, a weight matrix $\mW \in \mathbb{R}^{d \times k}$ is modified as:
% $$
% \mW' = \mW + \Delta \mW = \mW + \mB\mA,
% $$
% where $\mB \in \mathbb{R}^{d \times r}$ and $\mA \in \mathbb{R}^{r \times k}$ are the trainable low-rank matrices. \todoblue{can use the actual version that includes the hyperparameter $\alpha$ to be more accurate, or when it is needed in later sections.} The parameter $r$, which controls the rank of the update, is typically set such that $r \ll \min(d, k)$, making the adaptation computationally efficient. Additionally, LoRA can be applied to only certain parts of the model—for example, to attention layers, MLP blocks, or both. This modularity allows fine-grained control over which components are adapted during finetuning. 

\subsection{Evaluation of the Finetuned Model}

After safety alignment finetuning is completed, we evaluate two aspects of the model: (1) safety, which is assessed using a dataset of harmful questions. We sample responses from the model for these questions and use \texttt{Llama-Guard-3-8B}—a model specialized in safety evaluation and shown to be the strongest safety evaluator in \citep{jiang2025safechain}—to determine whether the responses are safe. The \textit{safety score} is defined as the proportion of questions for which the model’s response is judged to be harmful. (2) reasoning ability, evaluated using multiple standard benchmark datasets containing questions on math, science, and coding—widely used to assess models' reasoning capabilities. We consider the commonly used metric Pass@1 to measure accuracy on these benchmarks. For each question, we sample $n$ responses, compute the fraction of correct responses, and then average this accuracy over all questions. We set $n=8$.

\subsection{Datasets and Models}

\textbf{Models.} We consider three widely used open-weight reasoning models: \texttt{DeepSeek-R1-Distill-Qwen-7B}, \texttt{DeepSeek-R1-Distill-Qwen-14B}, and \texttt{DeepSeek-R1-Distill-Llama-8B}. These models span three sizes and two architectures (Qwen and Llama). Safety evaluation is performed using \texttt{Llama-Guard-3-8B}, which \citet{jiang2025safechain} found to be the most accurate safety evaluator.

\textbf{Safety finetuning dataset.} We use the \texttt{DirectRefusal} dataset, adapted from \citep{rosati2024representation} by \citep{huang2025safety}, containing harmful requests with refusal answers. \looseness=-1

\textbf{Safety evaluation datasets.} We consider two datasets for safety evaluation.
The first is \texttt{StrongREJECT} \citep{souly2024strongreject}, which consists of 310 policy-violating queries designed to test whether a model behaves safely.
The second is \texttt{BeaverTails} \citep{ji2023beavertails}, which contains harmful data spanning 14 harm categories.
Together, these datasets provide a comprehensive evaluation of model safety.
All results in the main paper are reported on \texttt{StrongREJECT}, and we defer results on \texttt{BeaverTails} to Appendix~\ref{apdx:beavertails}.
The patterns observed in our experiments are consistent across both datasets. We \textbf{do not} consider jailbreaking benchmarks, as they are out of scope.
Jailbreaking typically requires defenses beyond safety-alignment finetuning, such as input preprocessing \citep{jain2023baseline,wang2024defending,xie2023defending}, external detectors/classifiers \citep{alon2023detecting,sharma2025constitutional}, or inference-time interventions \citep{xu2024safedecoding,zou2023representation}. Accordingly, we only focus on the capabilities that safety alignment is intended to address.

\textbf{Reasoning benchmarks.} We evaluate the models performance on (1) American Invitational
Mathematics Examination 2024 (\texttt{AIME}), (2) \texttt{GPQA} \citep{rein2024gpqa} evaluate mathematical and scientific reasoning, respectively, (3) \texttt{HumanEval} \citep{chen2021evaluating} and (4) \texttt{MBPP} \citep{austin2021program} are code generation benchmarks. We also consider the augmented versions created by \texttt{EvalPlus} \citep{liu2023your}, denoted as \texttt{HumanEval+} and \texttt{MBPP+}. 

\textbf{Training Setup.} Full-model finetuning is performed for 5 epochs and LoRA is run for 10 epochs. We save and evaluate checkpoints at every epoch. Unless otherwise stated, LoRA is applied only to the MLP layers with rank $r\!=\!1$. In Section~\ref{sec: ablation}, we investigate the effect of varying $r$ and applying LoRA to different modules (e.g., MLP vs. attention) and layers. Additional details are deferred to Appendix \ref{app: exp_details}.\looseness=-1

\subsection{LoRA is All You Need for Safety Alignment of Reasoning LLMS}\label{sec: lora_bypass}

\begin{figure*}[!t]
    \centering
    % Left group (2x2)
    \begin{subfigure}[t]{0.32\textwidth}
        \centering
        \begin{subfigure}[t]{0.48\textwidth}
            \includegraphics[width=\linewidth]{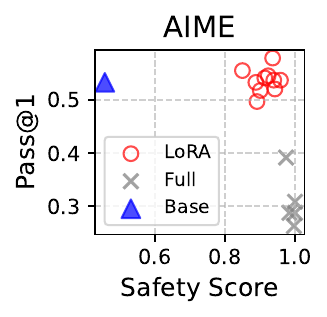}
        \end{subfigure}
        \begin{subfigure}[t]{0.48\textwidth}
            \includegraphics[width=\linewidth]{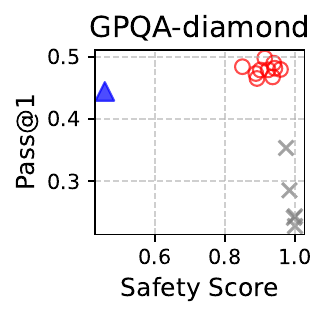}
        \end{subfigure}
        \begin{subfigure}[t]{0.48\textwidth}
            \includegraphics[width=\linewidth]{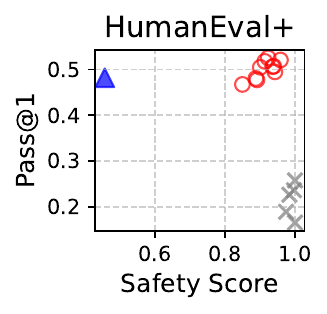}
        \end{subfigure}
        \begin{subfigure}[t]{0.48\textwidth}
            \includegraphics[width=\linewidth]{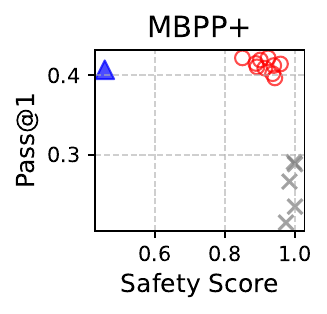}
        \end{subfigure}
        \caption{\texttt{R1-Distill-Qwen-7B}}
    \end{subfigure}
    \hfill
    % Right group (2x2)
    \begin{subfigure}[t]{0.32\textwidth}
        \centering
        \begin{subfigure}[t]{0.48\textwidth}
            \includegraphics[width=\linewidth]{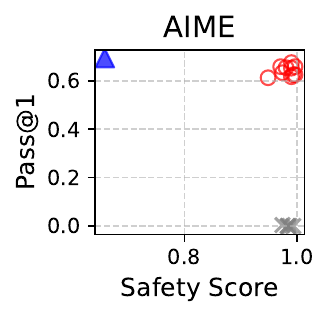}
        \end{subfigure}
        \begin{subfigure}[t]{0.48\textwidth}
            \includegraphics[width=\linewidth]{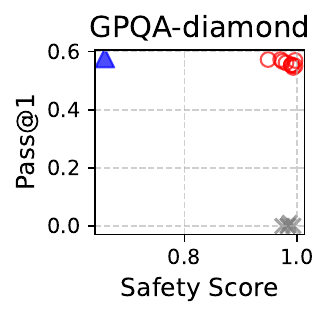}
        \end{subfigure}
        \begin{subfigure}[t]{0.48\textwidth}
            \includegraphics[width=\linewidth]{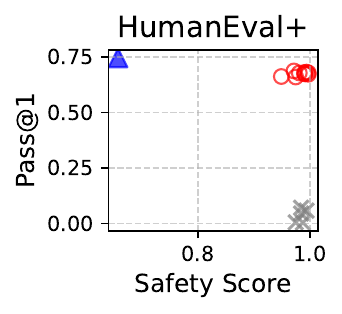}
        \end{subfigure}
        \begin{subfigure}[t]{0.48\textwidth}
            \includegraphics[width=\linewidth]{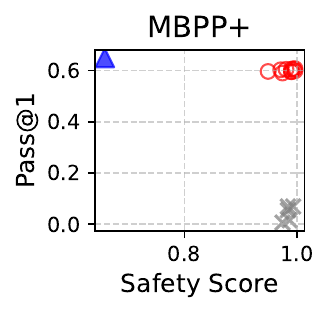}
        \end{subfigure}
        \caption{\texttt{R1-Distill-Qwen-14B}}
    \end{subfigure}
    \hfill
    \begin{subfigure}[t]{0.32\textwidth}
        \centering
        \begin{subfigure}[t]{0.48\textwidth}
            \includegraphics[width=\linewidth]{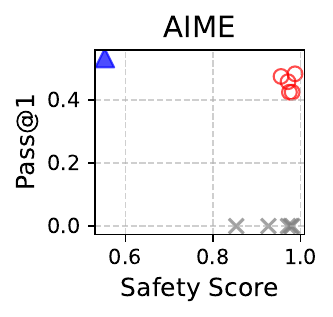}
        \end{subfigure}
        \begin{subfigure}[t]{0.48\textwidth}
            \includegraphics[width=\linewidth]{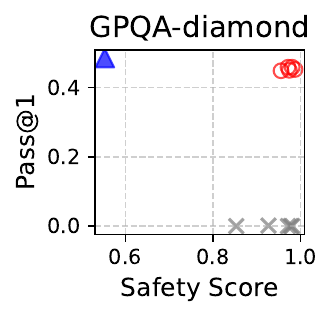}
        \end{subfigure}
        \begin{subfigure}[t]{0.48\textwidth}
            \includegraphics[width=\linewidth]{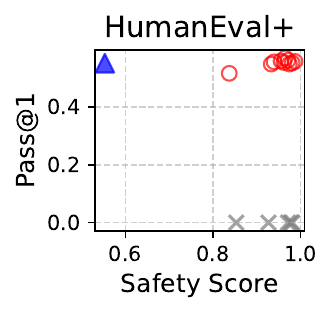}
        \end{subfigure}
        \begin{subfigure}[t]{0.48\textwidth}
            \includegraphics[width=\linewidth]{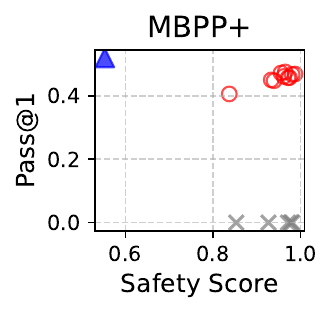}
        \end{subfigure}
        \caption{\texttt{R1-Distill-Llama-8B}}
    \end{subfigure}
    \caption{ Across different model sizes and architectures, LoRA bypasses the ``Safety Tax'', achieving safety comparable to full-model finetuning while preserving reasoning performance close to that of the original reasoning model. We plot reasoning performance—measured by Pass@1—against safety scores for different models. For finetuned models, we report results for checkpoints from all epochs. Results on the base versions of \texttt{HumanEval} and \texttt{MBPP} are shown in Figure~\ref{fig:lora_vs_full_code_base}; the same pattern holds, but with higher accuracy. All safety scores here are measured on \texttt{StrongREJECT}; results on \texttt{BeaverTails} are deferred to Figure~\ref{fig:beavertails}, where we observe the same pattern.}
    \label{fig:lora_vs_full}
    \vspace{-.2cm}
\end{figure*}

% In this experiment, we train both full-model and LoRA models for varying numbers of epochs and
Figure~\ref{fig:lora_vs_full} compares safety and reasoning performance across checkpoints (i.e., epochs) during full-model and LoRA-based safety alignment. We observe that the base model (before safety finetuning) achieves high accuracy but low safety. In contrast, full-model finetuning substantially improves safety at the cost of reduced reasoning accuracy. LoRA finetuning largely avoids this trade-off, maintaining strong performance on both safety and reasoning (as evidenced by the red points in the upper-right region of the plot). This pattern holds consistently across three model sizes, two architectures, two safety benchmarks, and four reasoning benchmarks spanning mathematics, science, and code generation. In some settings—e.g., the 14B model on the two coding tasks or the 8B model on \texttt{AIME}—we observe a slight drop in reasoning performance relative to the base model. In other cases, LoRA finetuning even improves reasoning performance (e.g., the 7B model across tasks). Overall, the trend is robust and highlights the effectiveness of LoRA compared to full-model finetuning.

\section{How Much LoRA is Enough?}\label{sec: ablation}

In this section, we ablate the LoRA configuration to identify the key elements that matter most. We examine three factors: (1) the rank \( r \), (2) the modules and (3) the layers to which LoRA is applied. Our goal is to determine the minimal setup needed for LoRA—i.e., the smallest update sufficient to achieve both strong reasoning and safety.

To compare LoRA configurations, we visualize performance in the reasoning–safety plane. For each configuration, we train for 10 epochs, producing multiple checkpoints with varying reasoning and safety scores. We evaluate each configuration via the \emph{Pareto frontier} of its checkpoints, enabling a clean visualization and effective comparison. \looseness=-1

\subsection{Rank: $r=1$ is Sufficient}
\begin{figure}[!t]
    \centering
    \vspace{-.3cm}
        \includegraphics[width=.85\linewidth]{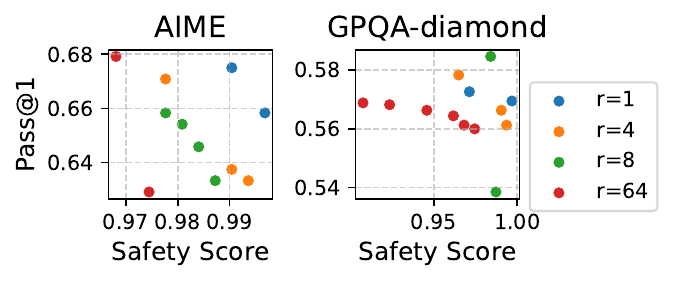}
    \vspace{-.12cm}
    \caption{ We visualize the Pareto frontiers of the reasoning–safety tradeoff when the training epoch is varied and observe that \( r=1 \) is sufficient to achieve an excellent tradeoff, outperforming other ranks, especially on AIME.}
    \label{fig:r_14b}
    \vspace{-.4cm}
\end{figure}

We explore the effect of the LoRA rank \( r \) by applying LoRA to the MLP layers of the 14B model while varying \( r \).
We find that \( r = 1 \) is sufficient to achieve the best performance in both reasoning and safety.
As shown in Figure~\ref{fig:r_14b}, \( r = 1 \) attains the best (on AIME) or near-best (on GPQA) trade-off across training epochs.
This result suggests that the safety alignment task is inherently low-rank.
Indeed, prior work has shown that safety behavior can often be mediated by a single direction in a model's internal representations,
commonly referred to as a steering vector \citep{panickssery2023steering} or refusal features
\citep{arditi2024refusal,yu2024robust}, which may explain why a rank-1 update suffices for effective safety alignment.

\subsection{Modules: Up-Projection Matters Most in MLPs}

\begin{figure}[!t]
\vspace{-.2cm}
    \centering
    \includegraphics[width=0.34\textwidth]{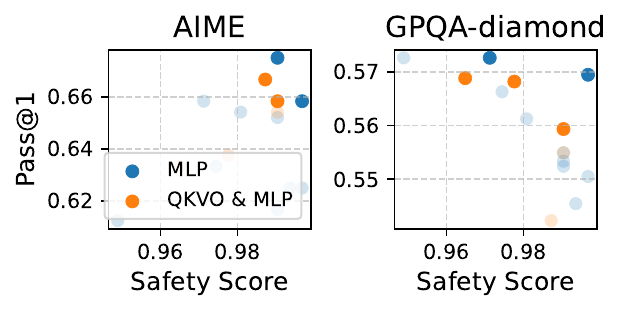}
    \vspace{-.2cm}
    \caption{Applying LoRA to MLP modules alone is sufficient. Faded points indicate non–Pareto-frontier points.}
    \label{fig:qkvo_mlp}
    \vspace{-.4cm}
\end{figure}

\begin{figure*}[!t]
    \centering
\includegraphics[width=.42\linewidth]{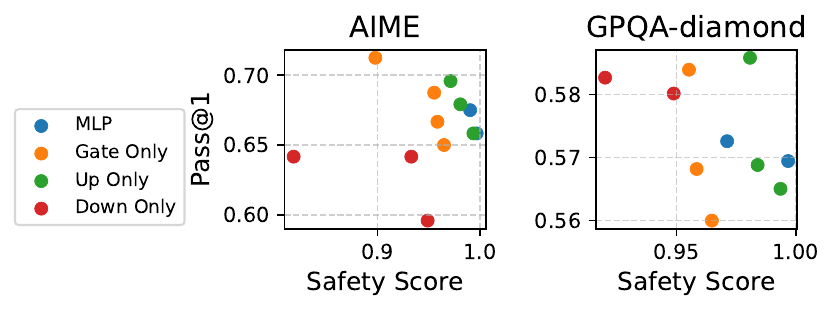}
\hspace{1cm}
\includegraphics[width=.32\linewidth]{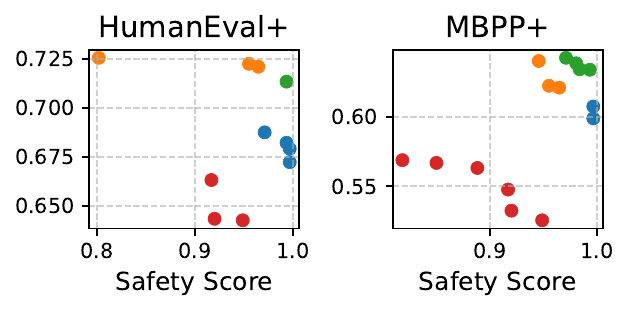}    
    \caption{ We compare applying LoRA to different projections within the MLP layers. The results show that applying it only to the up projection achieves the best tradeoff, and even outperforms applying it to the full MLP on the coding benchmarks. }
    \label{fig:modules}
\end{figure*}

\begin{figure*}[!t]
    \centering
    \includegraphics[width=0.42\linewidth]{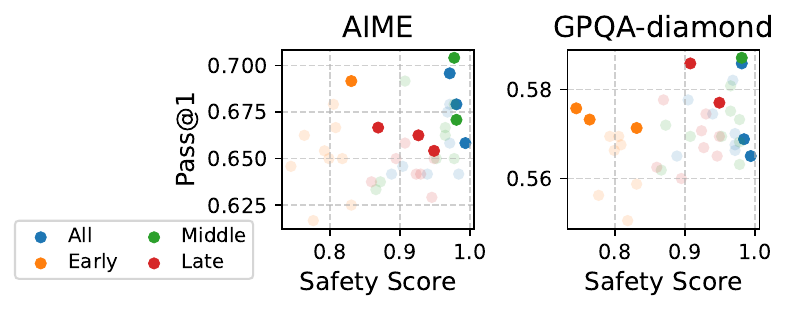}
    \hspace{1cm}
    \includegraphics[width=0.32\linewidth]{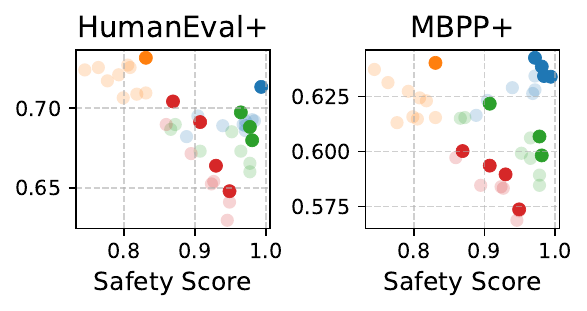}
    \caption{Applying LoRA to the middle layers (17–32) achieves a better tradeoff compared to using either the early or late layers, and performs on par with or only slightly behind using all layers across tasks. In the plot, the faded points indicate non–Pareto-frontier points.}
    \label{fig:layers}
    \vspace{-.2cm}
\end{figure*}

The LoRA adapter is usually applied to attention layers and/or MLP modules, e.g., by setting \texttt{target\_modules} when using the \texttt{PEFT} package. Here, we explore the effect of applying LoRA to different modules. We first compare applying it to both attention and MLP layers (QKVO \& MLP in the figure) versus only applying it to MLP layers. Figure~\ref{fig:qkvo_mlp} shows the comparison for the 14B models with $r=1$. We observe that applying LoRA only to the MLP layers yields a similar Pareto frontier compared to applying it to both.\looseness=-1

Next, we perform an ablation over the modules within the MLP. Specifically, in the \texttt{Qwen} architecture, the MLP layers use the popular SwiGLU \cite{chowdhery2022palm} structure that contains a gate projection, an up projection, and a down projection. We apply LoRA to only one of these projection layers at a time and evaluate the results. We use $r=1$. Figure~\ref{fig:modules} shows that across different tasks, applying LoRA only to the up projection achieves strong results, with the Pareto frontier often on par with using the full MLP and even outperforming it on the two coding benchmarks. In contrast, applying LoRA only to the down projection yields noticeably worse performance. These findings suggest that different projections within the MLP contribute differently to the reasoning–safety tradeoff, and that the up projection is particularly important and sufficient by itself in our setup.

\textbf{Discussion.} We mainly focus our ablations on the MLP, as it is primarily responsible for feature transformations and thus well-suited for safety alignment. We believe it is also worthwhile to study ablations over the attention layers, which we leave for future work. It is also valuable to further investigate why the up projection alone is the best choice. Additionally, extending these experiments to other LLM architectures would be an important direction for future work.\looseness=-1

\subsection{Layers: Middle Layers Matter Most}

We ablate over layers in the model. The 14B \texttt{Qwen2.5} architecture has 48 layers in total, and we apply LoRA to only 16 of them. We consider three configurations: (1) layers 5–20, denoted as ``Early Layers'', (2) layers 17–32, ``Middle Layers'', and (3) layers 25–40, ``Late Layers''. In all cases, we apply LoRA only to the up projection layers with $r=1$. The results are shown in Figure~\ref{fig:layers}. Across tasks, we observe that applying LoRA to the middle layers yields the best tradeoff, achieving performance on par with using all layers on \texttt{AIME} and \texttt{GPQA}, and only slightly behind using all layers on \texttt{HumanEval+} and \texttt{MBPP+}. In contrast, applying LoRA to either the early or late layers results in a noticeably worse tradeoff. This shows that the middle layers are most important for balancing reasoning and safety. Interestingly, this again connects to prior findings on steering vectors \citep{panickssery2023steering} and refusal features \citep{arditi2024refusal,yu2024robust}, which suggest that directions in intermediate representations responsible for safety behavior are most prominent in the middle layers.\looseness=-1

\section{A Theoretical Analysis}

We present a theoretical analysis that isolates the key tension in finetuning: adapting to a new task while preserving performance on a base task. We show that full-model finetuning fits the finetuning task perfectly but can catastrophically forget the base task, whereas LoRA—by restricting updates to be low-rank—can simultaneously fit the finetuning task and retain the base task when the finetuning target is low-rank and the base target has large intrinsic dimension. Moreover, the analysis reveals that a higher-dimensional base task is more tolerant to rank overshoot, providing insight into when and why LoRA works in practice.

\subsection{Setup}
\label{sec:theory_linear}

\textbf{Data and model.}
We study a family of regression tasks where the goal is to learn a linear map from $\sR^d$ to $\sR^k$.
Each task is specified by $\gT=(\gV,s,\sigma,\mTheta)$, where $\gV\subseteq\sR^d$ is a task-specific signal subspace,
$s>0$ is the signal scale, $\sigma>0$ is the noise scale, and $\mTheta\in\sR^{d\times k}$ is the target map.
Data are generated by $\vx=\vz+\vxi$ with $\vz\sim\cN(0,s^2\mP_{\gV})$ and $\vxi\sim\cN(0,\sigma^2\mP_{\gV^\perp})$, and labels are $\vy=\vz^\top\mTheta$,
where $\mP_{\gV}\mTheta=\mTheta$.
We consider linear predictors $f_{\mW}(\vx)=\vx^\top\mW$ trained under the squared loss $\displaystyle
L(\mW,\gT)\!=\!\E\bigl[\|\vx^\top\mW\!-\!\vy\|^2\bigr]$. For comparability across tasks, we fix the label scale such that \( \E\|\vy\|^2 = C \) for a fixed constant \( C > 0 \).

\textbf{Base vs.\ finetuning tasks.}
We consider a base task $\gT_{\rm base}=(\gV_{\rm base},s_{\rm base},\sigma_{\rm base},\mTheta_{\rm base})$
and a finetuning task $\gT_{\rm ft}=(\gV_{\rm ft},s_{\rm ft},\sigma_{\rm ft},\mTheta_{\rm ft})$, and assume the following:
\begin{enumerate}[label=(A\arabic*)]
\vspace{-.2cm}
\item \label{as:V_orth_main}
$\gV_{\rm base}\perp \gV_{\rm ft}$, i.e., the signal subspaces of the base and finetuning tasks are orthogonal.
\vspace{-.2cm}
\item \label{as:noise_sep_main}
$0<\!\sigma_{\rm base}\!=\!o(s_{\rm ft})$ and $0\!<\!\sigma_{\rm ft}\!=\!o(s_{\rm base})$, i.e., each task has nonzero noise, but its noise is asymptotically much smaller than the other task’s signal.
\vspace{-.2cm}
\end{enumerate}
\noindent
Additionally, the following spectral separation condition is used only for Theorem~\ref{thm:lora_tradeoff_main}:
\begin{enumerate}[label=(A\arabic*),resume]
\item \label{as:spectral_sep_main}
$
\frac{\sigma_1(\mTheta_{\rm base})}{\sigma_{r_{\rm ft}}(\mTheta_{\rm ft})}
<
\frac{s_{\rm ft}}{\sigma_{\rm ft}}
$, where $\sigma_i(\cdot)$ denotes the $i$-th singular value in nonincreasing order.
In words, the ratio between the largest singular value of $\mTheta_{\rm base}$ and the smallest nonzero singular value of $\mTheta_{\rm ft}$
is strictly smaller than the signal-to-noise ratio of the finetuning task.
\end{enumerate}

\textbf{Finetuning from a base model.} Finetuning initializes at $\mW_0=\mTheta_{\rm base}$, which is optimal for the base task, and then finetunes the model on the finetuning task.
We compare the following approaches. (i) \emph{full-model finetuning}. We run gradient descent over $\mW$ to minimize $L(\mW,\gT_{\rm ft})$. Since the population loss is a convex quadratic function of $\mW$ with a unique minimizer, classical results in convex optimization imply that, for a sufficiently small stepsize $
0<\eta<\frac{1}{\max\{s_{\mathrm{ft}}^2,\sigma_{\mathrm{ft}}^2\}}$, gradient descent converges linearly to the unique minimizer.
We therefore focus on the converged solution, which we denote by\looseness=-1
$$
\mW_{\mathrm{full}}
\in
\arg\min_{\mW} L(\mW,\gT_{\mathrm{ft}}),
$$
and treat it as the final model obtained by full-model finetuning. (ii) \emph{LoRA finetuning.} We reparameterize the model weights as $\mW=\mW_0+\mB\mA$, where $\mB\in\sR^{d\times r}$ and $\mA\in\sR^{r\times k}$ are trainable parameters,
and perform gradient descent over $(\mB,\mA)$:
\[
(\mB_{t+1},\mA_{t+1})=(\mB_t,\mA_t)-\eta\,\nabla_{\mB,\mA}L(\mW_0+\mB_t\mA_t,\gT_{\mathrm{ft}}).
\]
Under mild conditions on the stepsize and initialization (see detailed discussion in Appendix \ref{apdx:gd}),
gradient descent converges to a global minimizer. We therefore focus on the converged parameters $(\mB_\infty,\mA_\infty)$. Equivalently, letting $\mS:=\mB\mA$, LoRA restricts the update to satisfy $\rank{\mS}\le r$.
Thus, the final weights obtained by LoRA can be written as\looseness=-1
\[
\mW_{\mathrm{lora}}=\mW_0+\mS^*,
\text{where}\,
\mS^*\in \arg\min_{\rank{\mS}\le r} L(\mW_0+\mS,\gT_{\mathrm{ft}}).
\]

\textbf{Key dimensions.}
We write $r_{\rm ft}:=\rank(\mTheta_{\rm ft})$ and define the \emph{intrinsic dimension} (stable rank)
\[
\intdim(\mM)\;:=\;\|\mM\|_F^2/\|\mM\|_2^2\;\in[1,\rank(\mM)].
\]

\textbf{Overshoot.} We also define the \emph{overshoot}
\[
\Delta \;:=\; (r-r_{\rm ft})_+,
\]
measuring how much the LoRA budget exceeds the minimal rank needed to represent the finetuning map exactly.\looseness=-1

\subsection{Comparison and a Deep Dive into LoRA}

\begin{figure}[!t]
\vspace{-.2cm}
    \centering
    \includegraphics[width=0.49\textwidth]{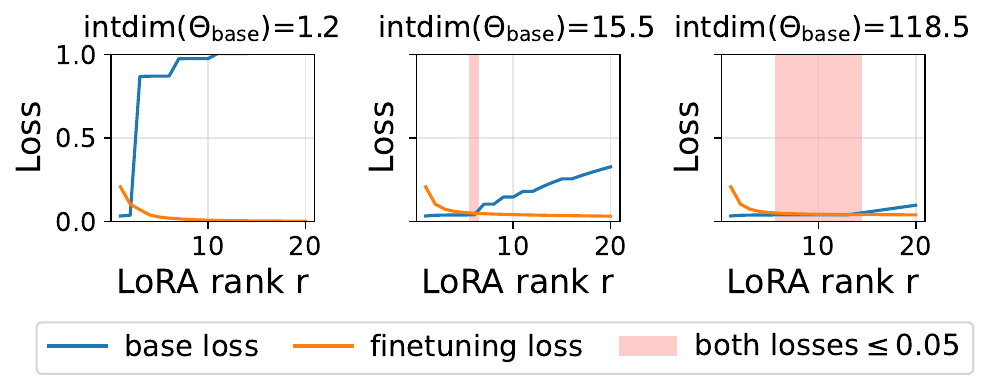}
    \vspace{-.3cm}
\caption{
Numerical experiments under the theoretical setup. We let the singular values of \( \mTheta_{\rm base} \) and \( \mTheta_{\rm ft} \) follow a power-law decay. We plot both losses against the LoRA rank \( r \).
\( \intdim(\mTheta_{\rm ft}) \) is fixed at \( 1.2 \),
while \( \intdim(\mTheta_{\rm base}) \) increases from left to right
(\( 1.2 \), \( 15.5 \), and \( 118.5 \)).
The red shaded region indicates values of \( r \) for which both losses are \(\leq\! 0.05 \).
As \( \intdim(\mTheta_{\rm base}) \) grows, this region expands from nonexistent to wide.\looseness=-1
%When $\intdim(\mTheta_{\rm base})$ is small (left), there is no $r$ that achieves low loss on both tasks.
% For an intermediate intrinsic dimension (middle), good performance is achievable but highly sensitive to overshoot, yielding a narrow feasible region.
% When $\intdim(\mTheta_{\rm base})$ is large (right), the feasible region widens substantially, illustrating increased tolerance to overshoot and reduced sensitivity to the precise choice of $r$.
}
    \label{fig:good_region}
    \vspace{-.2cm}
\end{figure}

\textbf{Full-model finetuning forgets.}
Full finetuning achieves the finetuning optimum but overwrites the base solution.

\begin{theorem}[Full-model finetuning forgets]
\label{thm:full_forgets_main}
Assume \ref{as:V_orth_main}--\ref{as:noise_sep_main}. 
As proved in Appendix~\ref{subsec:proof_full_forgets}, full finetuning satisfies\looseness=-1
\[
L(\mW_{\rm full},\gT_{\rm ft})=0,
\qquad
L(\mW_{\rm full},\gT_{\rm base})\ge C.
\]
\end{theorem}

\textbf{The finetuning task must be low-dimensional for LoRA to work.}
Even ignoring interference with the base task, a rank-$r$ update can capture at most a rank-$r$ approximation of $\mTheta_{\rm ft}$.
The next result shows that the finetuning loss cannot be small unless the intrinsic dimension of $\mTheta_{\rm ft}$ is comparable to the chosen rank $r$. The proof is in Appendix \ref{subsec:proof_ft_lb}.
\begin{theorem}
\label{thm:lora_ft_lb_main}
Assume \ref{as:V_orth_main}.
For any LoRA solution with $r$, \looseness=-1
\[
L(\mW_{\rm lora},\gT_{\rm ft})
\;\ge\;
C\Bigl(1-\frac{r}{\intdim(\mTheta_{\rm ft})}\Bigr)_+.
\]
Consequently, if $\intdim(\mTheta_{\rm ft})\!\gg \!r$, then $L(\mW_{\rm lora},\gT_{\rm ft})\gtrsim C$.\looseness=-1
\end{theorem}

\begin{figure}[!t]
    \centering
    \includegraphics[width=0.32\linewidth]{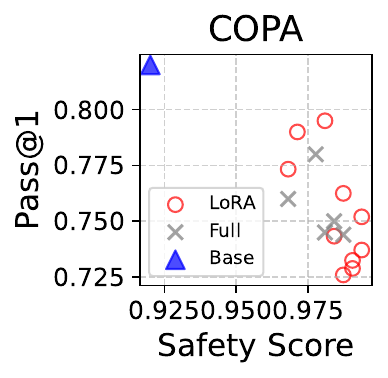}
    \includegraphics[width=0.32\linewidth]{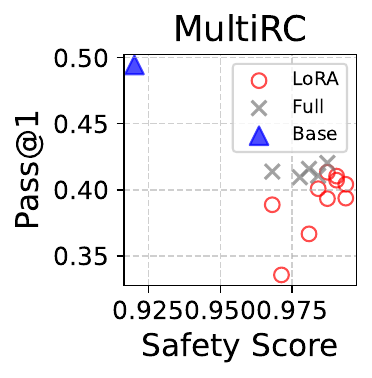}
    \includegraphics[width=0.32\linewidth]{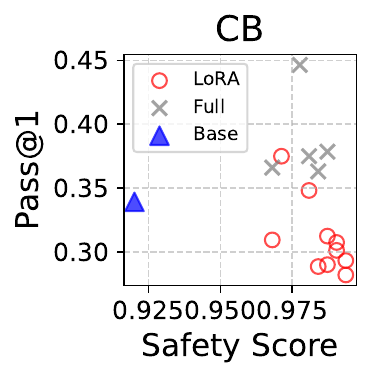}
    \caption{
Safety finetuning on \texttt{Qwen2-1.5B-Instruct}.
Utility is evaluated on instruction-following tasks (\texttt{COPA}, \texttt{MultiRC}, \texttt{CB}), and safety on \texttt{StrongREJECT}.
In this non-reasoning, instruction-following setting, LoRA no longer provide an effective safety--utility trade-off.
}
    \label{fig:non_reasoning}
    \vspace{-.2cm}
\end{figure}

\textbf{Interaction between overshoot and the intrinsic dimension of the base task.}
When $r$ is large enough to represent the finetuning map, LoRA can fit the finetuning task while approximately preserving the base task.
However, the degree of base-task degradation depends on the overshoot $\Delta$ and on how spread out the base task is in parameter space.
A large $\intdim(\mTheta_{\rm base})$ means the energy of $\mTheta_{\rm base}$ is distributed across many directions,
making the base task more tolerant to overshoot.
The next result formalizes this: once $r$ is large enough to fit the finetuning map, the finetuning loss becomes negligible,
and the base loss scales inversely with $\intdim(\mTheta_{\rm base})$. The proof is in Appendix \ref{subsec:proof_tradeoff}.

\begin{theorem}[Near-perfect LoRA finetuning when the base task has large intrinsic dimension]
\label{thm:lora_tradeoff_main}
Assume \ref{as:V_orth_main}--\ref{as:spectral_sep_main}.
If $r\ge r_{\rm ft}$, then the LoRA-finetuned model satisfies
\begin{align*}
L(\mW_{\rm lora},\gT_{\rm ft}) &= o(1),\\
L(\mW_{\rm lora},\gT_{\rm base})
&\le
C\cdot\frac{\Delta}{\intdim(\mTheta_{\rm base})}+o(1).
\end{align*}
Consequently, if $\intdim(\mTheta_{\rm base})=\omega(\Delta)$, then $L(\mW_{\rm lora},\gT_{\rm base})=o(1)$ even with overshoot.
\end{theorem}

\begin{remark}[Lower bounds and tightness]
\label{rem:base_lb} The following rank-based lower bound always holds:\looseness=-1
\[
L(\mW_{\rm lora},\gT_{\rm base})
\;\ge\;
C\cdot\frac{\Delta}{\rank(\mTheta_{\rm base})}+o(1).
\]
Moreover, under a mild flat-spectrum condition (see Appendix~\ref{subsec:base_lb}),
one obtains a matching lower bound of order\looseness=-1
\[
L(\mW_{\rm lora},\gT_{\rm base})
\;\gtrsim\;
C\,\frac{\Delta}{\intdim(\mTheta_{\rm base})}.
\]
\end{remark}
Importantly, Theorem~\ref{thm:lora_tradeoff_main} and Remark~\ref{rem:base_lb} show that the sensitivity to overshoot $\Delta$ is governed by
$\intdim(\mTheta_{\rm base})$.
In practice, it is rarely possible to choose the LoRA rank $r$ perfectly, and since $r$ is discrete, any mismatch yields $\Delta\ge 1$.
Therefore, a key requirement for robust performance is that the base task have sufficiently large intrinsic dimension, so that it can tolerate inevitable overshoot. 

To illustrate this effect, we conduct numerical experiments that mirror our theoretical setup, setting the singular values of $\mTheta_{\rm base}$ and $\mTheta_{\rm ft}$ to follow a power-law decay. Fig.~\ref{fig:good_region} plots the base and finetuning losses as a function of the LoRA rank $r$ across three regimes, with the intrinsic dimension of the base task increasing from left to right. In the leftmost panel ($\intdim(\mTheta_{\rm base})\!=\!1.2$), no choice of $r$ achieves good performance on both tasks. In the middle panel ($\intdim(\mTheta_{\rm base})\!=\!15.5$), good performance is achievable, but the base loss is highly sensitive and increases sharply even with a small overshoot; correspondingly, the red ``good'' region is narrow and requires a very precise choice of $r$. In the rightmost panel ($\intdim(\mTheta_{\rm base})\!=\!118.5$), the good region widens substantially, demonstrating increased tolerance to overshoot and making LoRA much less sensitive to the exact choice of $r$.\looseness=-1

\textbf{Taken together: LoRA succeeds with high-rank base tasks and low-rank finetuning tasks.}
Theorems~\ref{thm:lora_ft_lb_main} and \ref{thm:lora_tradeoff_main} formalize two complementary conditions under which LoRA is effective:
the finetuning task must be sufficiently low-rank (otherwise the finetuning loss is large),
and the base task must be sufficiently high-rank to tolerate the inevitable overshoot encountered in practice. This perspective helps explain why LoRA is effective in our earlier experiments.
In particular, safety finetuning a reasoning model matches this regime: the base capability we aim to preserve---\emph{reasoning}---is highly complex and thus plausibly ``high-rank'' in nature.
In contrast, the finetuning objective---\emph{safety}---appears to be relatively low-rank, as suggested by prior work on steering vectors \citep{panickssery2023steering}
and ``refusal'' features \citep{arditi2024refusal,yu2024robust}, which show that safety-related behavior can be mediated by a single direction or a small number of directions in a model's internal representations.

To probe this prediction in a more realistic setting, we consider safety finetuning an instruction-tuned (non-reasoning) model, where the base capability is instruction following rather than reasoning.
We finetune \texttt{Qwen2-1.5B-Instruct} for safety and evaluate utility on instruction-following tasks \cite{wang2019superglue} including \texttt{COPA}, \texttt{MultiRC} and \texttt{CB}, and safety on \texttt{StrongREJECT}.
Fig.~\ref{fig:non_reasoning} shows that LoRA is no longer effective in this regime and fails to provide a favorable safety--utility trade-off.
This aligns with our theoretical insight that LoRA performs best when the base capability to be preserved is intrinsically high-dimensional.
Here, the relevant base capability is instruction following, which is intuitively less complex than reasoning and thus plausibly lower-dimensional, potentially accounting for LoRA's weaker performance in this setting.\looseness=-1

\section{Conclusion}

Applying LoRA during safety alignment of reasoning LLMs can effectively mitigate the previously observed ``Safety Tax'', achieving strong safety without compromising reasoning, as validated by our extensive experiments. We further ablate LoRA configurations and find that (1) rank-$1$ updates are sufficient and often optimal, (2) the MLP up-projection layers are the most important, and (3) middle layers matter more than early or late layers, providing practical guidance for achieving strong performance at minimal fine-tuning cost. Finally, our theoretical analysis clarifies when and why LoRA works, revealing novel insights into how rank overshoot interacts with the intrinsic dimensionality of the base capability.

\section*{Acknowledgement}
This research was supported in part by in part by the NSF CAREER Award 2146492, NSF-Simons AI Institute for Cosmic Origins (CosmicAI), and NSF AI Institute for Foundations of Machine Learning (IFML).

\bibliography{reference}
\bibliographystyle{icml2026}

%%%%%%%%%%%%%%%%%%%%%%%%%%%%%%%%%%%%%%%%%%%%%%%%%%%%%%%%%%%%%%%%%%%%%%%%%%%%%%%
%%%%%%%%%%%%%%%%%%%%%%%%%%%%%%%%%%%%%%%%%%%%%%%%%%%%%%%%%%%%%%%%%%%%%%%%%%%%%%%
% APPENDIX
%%%%%%%%%%%%%%%%%%%%%%%%%%%%%%%%%%%%%%%%%%%%%%%%%%%%%%%%%%%%%%%%%%%%%%%%%%%%%%%
%%%%%%%%%%%%%%%%%%%%%%%%%%%%%%%%%%%%%%%%%%%%%%%%%%%%%%%%%%%%%%%%%%%%%%%%%%%%%%%
\newpage
\appendix
\onecolumn

\section{Experimental Details}\label{app: exp_details}

\subsection{Details about \texttt{DirectRefusal}}

The original \texttt{DirectRefusal} dataset can be downloaded from \texttt{TianshengHuang/DirectRefusal} on Hugging Face. For each harmful request, the corresponding response includes a short thought—``I should not answer this question!”—followed by a refusal answer. We adapted the dataset to ensure it follows the \texttt{DeepSeek-R1} chat template.

\subsection{Details about Safety Evaluation}

For each question, we sample the answer using greedy decoding and pass the answer to \texttt{Llama-Guard-3-8B} to judge whether it is safe. Safety is determined by checking whether the substring ``safe" or ``unsafe" is contained in the evaluator's output.

\subsection{Details about \texttt{AIME} and \texttt{GPQA}}

We adapted the codebase from \cite{li2025small}, which builds on top of \texttt{lm-evaluation-harness} \cite{eval-harness}, to evaluate models on \texttt{GPQA} and \texttt{AIME}. For \texttt{AIME}, answers are evaluated using \texttt{Qwen2.5-32B-Instruct} as a judge. For \texttt{GPQA}, since it is a multiple-choice benchmark, answers are evaluated via regular expression matching. To improve evaluation accuracy, we made some small adjustments. One change is modifying the prompt to better ensure the model outputs answers in the desired format:
\begin{lstlisting}[basicstyle=\ttfamily\small, frame=single, breaklines=true]
You are solving a multiple-choice question. At the end, present your final answer using the format: Final Answer: \boxed{X}, where X is one of A, B, C, or D.

Question: {{Question}}
Choices:
(A) {{choice1}}
(B) {{choice2}}
(C) {{choice3}}
(D) {{choice4}}
\end{lstlisting}
Another adjustment is in the answer extraction logic—we enhanced the original implementation to handle a wider range of answer formats that models produce.

During generation, we set the temperature to 0.6, top\_p to 0.95, and the maximum number of generated tokens to 32{,}768.

\subsection{Details about \texttt{HumanEval} and \texttt{MBPP}}

We adapted the codebase of \texttt{EvalPlus} \cite{liu2023your}. The original implementation included a response prefix, designed for earlier models that did not explicitly support an intermediate reasoning process. This prefix—such as ``Below is a Python script with a self-contained function that efficiently solves the problem and passes corresponding tests:”—was prepended to model outputs during generation. However, this practice introduces unfair bias by encouraging all models to directly generate code. It inadvertently benefits the full-model fine-tuned baseline—which would otherwise often refuse to answer—by effectively forcing it to produce code. Conversely, it disadvantages reasoning-aligned models by disrupting the expected format that includes intermediate ``thought", causing them to skip the thinking process entirely. This can result in skewed conclusions. We suspect this explains the abnormally low accuracy of the base reasoning model and the high accuracy of the full-model fine-tuned variant reported in \cite{jiang2025safechain}. To address this issue, we remove the response prefix. We also slightly reword the prompt to better align with reasoning models.

During generation, we set the temperature to 0.6 and the maximum number of generated tokens to 32{,}768.

\subsection{Training Details}

For 7B models (Qwen) and 8B models (Llama), full-model finetuning is conducted using 4 GPUs with a batch size of 2 per device for 5 epochs. LoRA finetuning uses 2 GPUs with a batch size of 2 per device for 10 epochs. We set the LoRA hyperparameters as $\alpha = 16$ and \texttt{lora\_dropout} = 0.05.

For 14B models, full-model finetuning is performed using 8 GPUs with a batch size of 1 per device for 5 epochs. LoRA finetuning uses 4 GPUs with a batch size of 2 per device for 10 epochs. We set the LoRA hyperparameters as $\alpha = 16$ and \texttt{lora\_dropout} = 0.05.

Across all experiments, we use a learning rate of 5e-5 and a weight decay of 1e-4.

\section{Additional Plots for Experiments in Section~\ref{sec:lora_bypass}}

\subsection{Results on \texttt{BeaverTails}}\label{apdx:beavertails}

\begin{figure*}[!t]
    \centering
    % Left group (2x2)
    \begin{subfigure}[t]{0.32\textwidth}
        \centering
        \begin{subfigure}[t]{0.48\textwidth}
            \includegraphics[width=\linewidth]{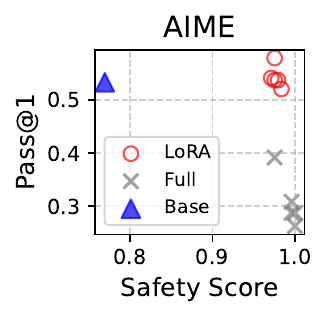}
        \end{subfigure}
        \begin{subfigure}[t]{0.48\textwidth}
            \includegraphics[width=\linewidth]{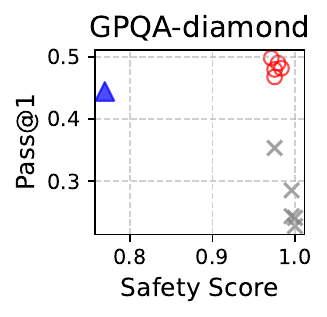}
        \end{subfigure}
        \begin{subfigure}[t]{0.48\textwidth}
            \includegraphics[width=\linewidth]{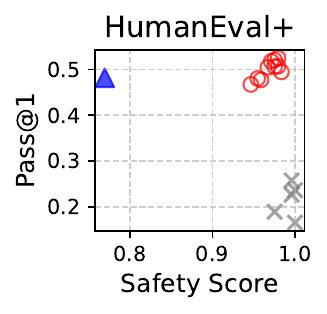}
        \end{subfigure}
        \begin{subfigure}[t]{0.48\textwidth}
            \includegraphics[width=\linewidth]{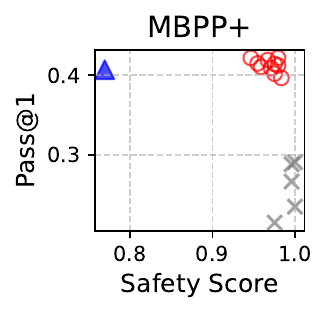}
        \end{subfigure}
        \caption{\texttt{R1-Distill-Qwen-7B}}
    \end{subfigure}
    \hfill
    % Right group (2x2)
    \begin{subfigure}[t]{0.32\textwidth}
        \centering
        \begin{subfigure}[t]{0.48\textwidth}
            \includegraphics[width=\linewidth]{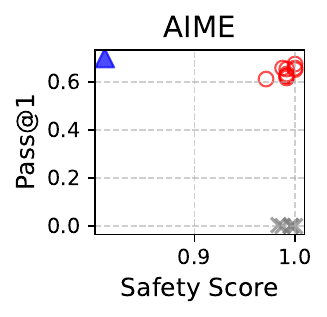}
        \end{subfigure}
        \begin{subfigure}[t]{0.48\textwidth}
            \includegraphics[width=\linewidth]{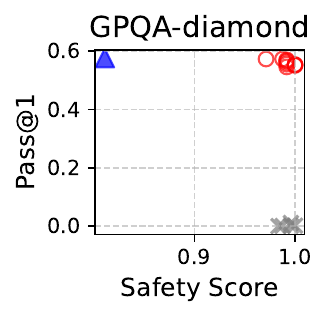}
        \end{subfigure}
        \begin{subfigure}[t]{0.48\textwidth}
            \includegraphics[width=\linewidth]{figures/safety_vs_acc/humaneval_14B_plus.pdf}
        \end{subfigure}
        \begin{subfigure}[t]{0.48\textwidth}
            \includegraphics[width=\linewidth]{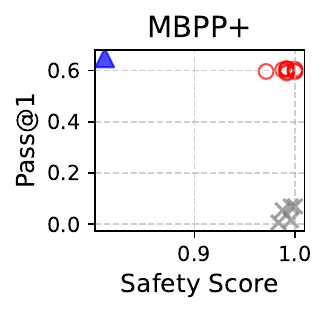}
        \end{subfigure}
        \caption{\texttt{R1-Distill-Qwen-14B}}
    \end{subfigure}
    \hfill
    \begin{subfigure}[t]{0.32\textwidth}
        \centering
        \begin{subfigure}[t]{0.48\textwidth}
            \includegraphics[width=\linewidth]{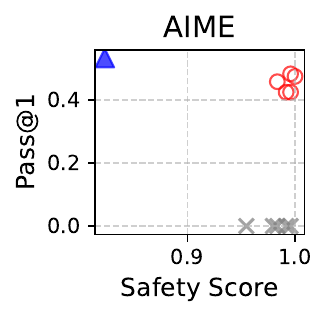}
        \end{subfigure}
        \begin{subfigure}[t]{0.48\textwidth}
            \includegraphics[width=\linewidth]{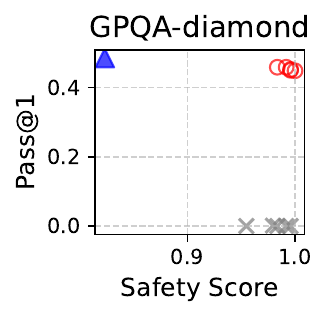}
        \end{subfigure}
        \begin{subfigure}[t]{0.48\textwidth}
            \includegraphics[width=\linewidth]{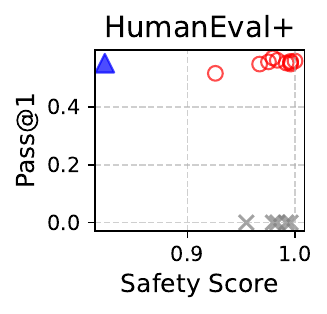}
        \end{subfigure}
        \begin{subfigure}[t]{0.48\textwidth}
            \includegraphics[width=\linewidth]{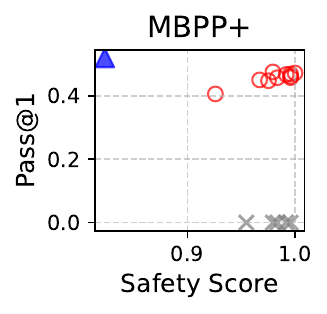}
        \end{subfigure}
        \caption{\texttt{R1-Distill-Llama-8B}}
    \end{subfigure}
    \caption{Results with safety measured on \texttt{BeaverTails}, consistent with the pattern shown in Figure~\ref{fig:lora_vs_full}.}
    \label{fig:beavertails}
\end{figure*}

The results on \texttt{BeaverTails} across models are shown in Figure~\ref{fig:beavertails}, and the observed pattern is consistent with that in Figure~\ref{fig:lora_vs_full}.

\subsection{Results on the Base Versions of \texttt{HumanEval} and \texttt{MBPP}}

\begin{figure}[!t]
    \centering
    % Left group (2x2)
    \begin{subfigure}[t]{0.32\textwidth}
        \centering

        \begin{subfigure}[t]{0.48\textwidth}
            \includegraphics[width=\linewidth]{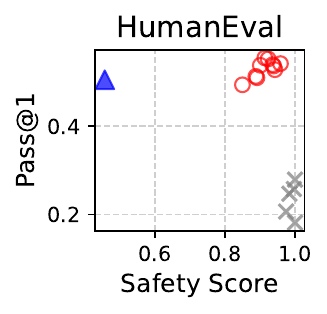}
        \end{subfigure}
        \begin{subfigure}[t]{0.48\textwidth}
            \includegraphics[width=\linewidth]{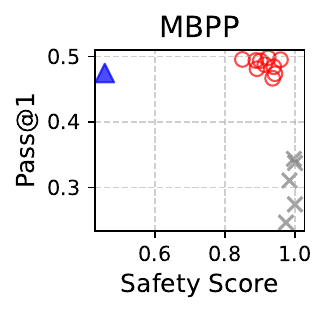}
        \end{subfigure}
        \caption{\texttt{R1-Distill-Qwen-7B}}
    \end{subfigure}
    \hfill
    % Right group (2x2)
    \begin{subfigure}[t]{0.32\textwidth}
        \centering
        \begin{subfigure}[t]{0.48\textwidth}
            \includegraphics[width=\linewidth]{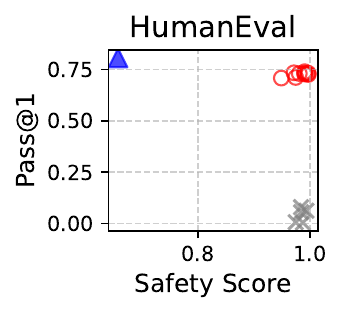}
        \end{subfigure}
        \begin{subfigure}[t]{0.48\textwidth}
            \includegraphics[width=\linewidth]{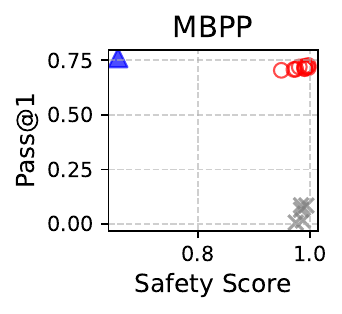}
        \end{subfigure}
        \caption{\texttt{R1-Distill-Qwen-14B}}
    \end{subfigure}
    \begin{subfigure}[t]{0.32\textwidth}
        \centering
        \begin{subfigure}[t]{0.48\textwidth}
            \includegraphics[width=\linewidth]{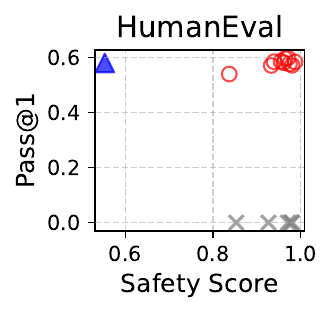}
        \end{subfigure}
        \begin{subfigure}[t]{0.48\textwidth}
            \includegraphics[width=\linewidth]{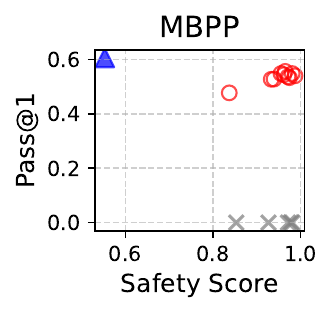}
        \end{subfigure}
        \caption{\texttt{R1-Distill-Llama-8B}}
    \end{subfigure}
    \caption{Results on the base versions of \texttt{HumanEval} and \texttt{MBPP} show the same pattern as in the plus versions shown in Figure~\ref{fig:lora_vs_full}, but with higher accuracy.}
    \label{fig:lora_vs_full_code_base}
\end{figure}

Figure \ref{fig:lora_vs_full_code_base}
shows the results on the base versions of \texttt{HumanEval} and \texttt{MBPP}.

\section{Additional Experimental Results: Exploring the Structure of LoRA Weights}

We empirically study the structure of LoRA weight updates. Preliminary results suggest that LoRA induces updates with slightly smaller overlap with the initial weights than full-model finetuning, which may indicate that LoRA is less disruptive to reasoning-related weights. We then explore methods to explicitly reduce this overlap (via regularization or post-hoc weight merging). One post-hoc method yields a modest improvement in the reasoning–safety trade-off on some tasks. These results are preliminary, and further work is needed to develop approaches that produce consistent gains across tasks; we hope our findings will inspire future work in this direction.

\begin{figure}[!t]
    \centering
    \begin{subfigure}[t]{0.24\textwidth}
        \centering
            \includegraphics[width=\linewidth]{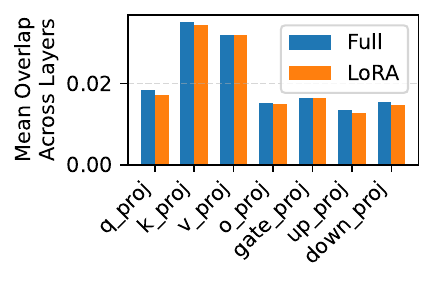}
        \caption{$ \frac{\| \mW_I^\top \Delta\mW \|}{\| \mW_I \| \| \Delta \mW \|} $}
        \label{fig:mean_alignment_col_prod}
    \end{subfigure}
    \begin{subfigure}[t]{0.24\textwidth}
        \centering
            \includegraphics[width=\linewidth]{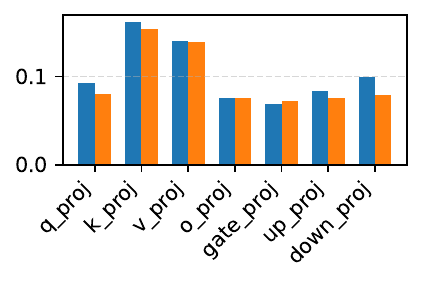}
        \caption{$\frac{ \| \mU_{16}\mU_{16}^\top\Delta\mW \| }{ \| \Delta\mW \|} $}
        \label{fig:mean_alignment_col_v}
    \end{subfigure}
    \begin{subfigure}[t]{0.24\textwidth}
        \centering
            \includegraphics[width=\linewidth]{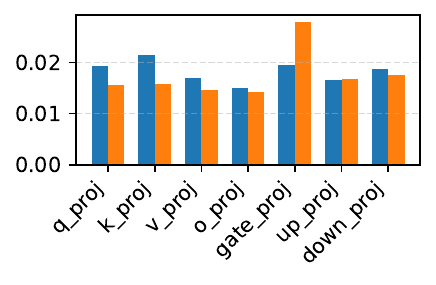}
        \caption{$ \frac{\| \mW_I \Delta\mW^\top \|}{\| \mW_I \| \| \Delta \mW \|} $}
        \label{fig:mean_alignment_row_prod}
    \end{subfigure}
    \begin{subfigure}[t]{0.24\textwidth}
        \centering
            \includegraphics[width=\linewidth]{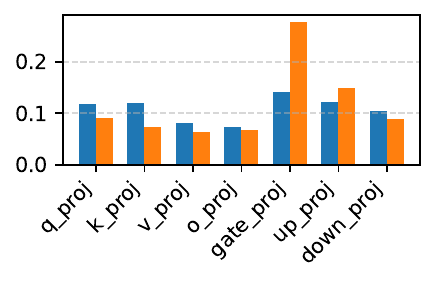}
        \caption{$\frac{ \| \mV_{16}\mV_{16}^\top\Delta\mW^\top \| }{ \| \Delta\mW \|} $}
        \label{fig:mean_alignment_row_v}
    \end{subfigure}
    \caption{LoRA updates exhibit smaller overlap with the original weights compared to the full-model finetuning updates. Although the reduction in overlap is sometimes small, it can be observed across most layers for all four metrics, which cover both the column (a)(b) and row spaces (c)(d). The 14B models are used here.  }
    \label{fig:mean_alignment_across_layers}
\end{figure}

\subsection{LoRA Updates Have Less Alignment with Initial Weights
}\label{sec: low_ortho}

Intuitively, if the initial (reasoning) weights $\mW_I$ and the LoRA update $\Delta\mW$ have only a small alignment, this suggests minimal interference between the safety-oriented update and the weights critical for reasoning. LoRA already constrains $\Delta\mW$ to be low-rank, meaning it spans only a small subspace of the full weight space. We further examine the orientation of this subspace: how do the directions learned by LoRA compare to those spanned by $\mW_I$? To quantify this, we compute the following four metrics: (1) $
\frac{\|\mW_I^\top \Delta\mW\|}{\| \mW_I \| \| \Delta\mW \|}$, (2) $\frac{ \| \mU_{16}\mU_{16}^\top\Delta\mW \| }{ \| \Delta\mW \|} $, (3) $\frac{\| \mW_I\Delta\mW^\top\|}{\| \mW_I \| \| \Delta\mW \|}$, and (4) $\frac{ \| \mV_{16}\mV_{16}^\top\Delta\mW^\top \| }{ \| \Delta\mW \|} $. Here, $\mU_{16}$ and $\mV_{16}$ are matrices containing the top 16 left and right singular vectors of $\mW_I$, respectively, obtained via truncated SVD. Intuitively, (1) and (2) capture the overlap between $\mW_I$ and $\Delta\mW$ in the column space: (1) is a matrix-level analogue of cosine similarity, while (2) measures the normalized projection of $\Delta\mW$ onto the top dominant directions of $\mW_I$. Similarly, (3) and (4) capture overlap in the row space. The column and row spaces correspond to the directions the matrices ``write to” and ``read from”, respectively. A smaller value in any of these metrics indicates greater orthogonality between $\mW_I$ and $\Delta\mW$ in the corresponding space.

We compare the full-rank fine-tuned model with the LoRA fine-tuned model in which both the attention and MLP modules are updated with $r=4$, making the two settings more comparable since updates occur in all major modules. We compute the alignment metrics for different module types across layers, average them over layers, and report the results for the 14B models in Figure~\ref{fig:mean_alignment_across_layers}. We observe that LoRA achieves smaller overlap in most modules across the metrics, with a few exceptions. This suggests that, for most weights, $\mW_I$ and $\Delta\mW$ are more orthogonal—both in the column and row spaces—for LoRA than for full-model finetuning. In other words, under LoRA finetuning, the safety-oriented updates read from and write into subspaces that are more separate from those used by the original reasoning-related weights—more so than in the full-model fine-tuned version. Although the reduction in alignment values is sometimes small, it may still indicate that LoRA updates interfere less with the reasoning-related components of the model, potentially explaining the better preservation of reasoning performance. A more in-depth investigation is needed to fully understand the underlying mechanisms and to develop more precise metrics for capturing this effect—an important direction for future work.

\looseness=-1

\begin{figure}[!t]
    \centering
    % Left group (2x2)
    % \begin{subfigure}[t]{0.48\textwidth}
    %     \centering
    %     \includegraphics[width=\linewidth]{figures/orthogonality/side_by_side_7B.pdf}
    %     \caption{7B models on \texttt{AIME} and \texttt{GPQA}.}
    %     \label{fig:ortho_math}
    % \end{subfigure}
    % \hfill
    % % Right group (2x2)
    % \begin{subfigure}[t]{0.48\textwidth}
    %     \centering
    %         \includegraphics[width=\linewidth]{figures/orthomerge/side_by_side_7B_plus.pdf}
    %     \caption{7B models on \texttt{HumanEval+} and \texttt{MBPP+}.}
    %     \label{fig:ortho_code}
    % \end{subfigure}

\begin{subfigure}[t]{0.42\textwidth}
        \centering
        \includegraphics[width=\linewidth]{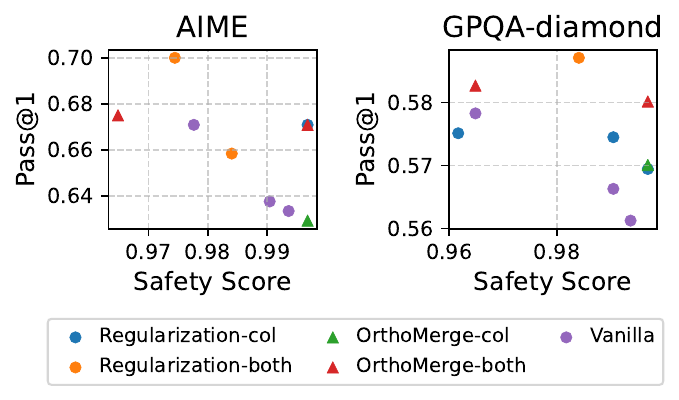}
        \caption{14B models on \texttt{AIME} and \texttt{GPQA}.}
        \label{fig:ortho_math_14b}
    \end{subfigure}
    \hspace{1cm}
    % Right group (2x2)
    \begin{subfigure}[t]{0.42\textwidth}
        \centering
            \includegraphics[width=\linewidth]{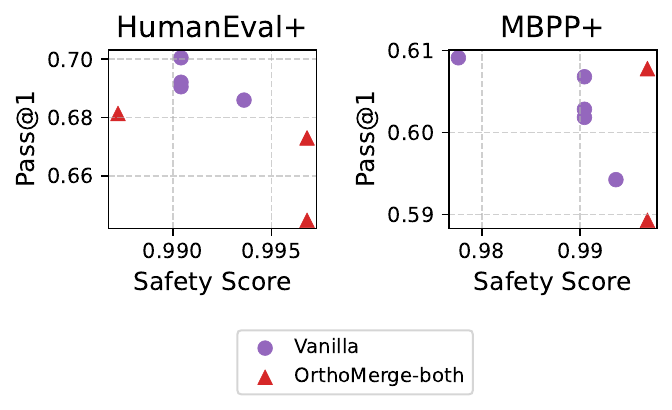}
        \caption{14B models on \texttt{HumanEval+} and \texttt{MBPP+}.}
        \label{fig:ortho_code_14b}
    \end{subfigure}
    
    \caption{ For each method that produces multiple checkpoints (e.g., across epochs or hyperparameter settings), we visualize the Pareto-frontier points. Methods enforcing orthogonality in both row and column spaces perform better than column-only variants. The post-hoc method \texttt{OrthoMerge-both} is the most promising, with points concentrated in the upper-right corner and its best point strictly dominating vanilla LoRA on \texttt{AIME} and \texttt{GPQA}. On \texttt{MBPP+}, it achieves a slightly better Pareto frontier, while on \texttt{HumanEval+} it slightly underperforms.
    }
    \label{fig:ortho}
\end{figure}

\subsection{Exploring Methods That Further Reduce Alignment}\label{subsec:try_methods}

Given the observations in Section~\ref{sec: low_ortho}, we ask whether further reducing the overlap between $\Delta\mW$ and $\mW_I$ could lead to even better reasoning performance without compromising safety. This question is particularly relevant because, while the results achieved by LoRA in Section~\ref{sec: lora_bypass} are strong, they are not perfect—a small performance gap remains compared to the original reasoning model, especially for the 14B model on \texttt{AIME}, \texttt{HumanEval+}, and \texttt{MBPP+}.

 We experiment with two approaches:  
    \begin{itemize}
        \item \textbf{Regularization during LoRA training.}   We add a penalty term to the loss that discourages overlap between $\mW_I$ and $\Delta\mW$ (to avoid out-of-memory issues during training caused by the large dimensionality of model weights, in implementation we use a low-rank approximation of $\mW_I$ instead of the full matrix), specifically:
    \begin{itemize}[leftmargin=0.8cm]
\setlength\itemsep{0em}
        \item \texttt{reg-col}: $\beta (\frac{\|\mW_I^\top \Delta\mW\|}{\| \mW_I \| \| \Delta\mW \|)})^2$ encourages orthogonality in the column space. 
        \item \texttt{reg-both}: $\beta (\frac{\|\mW_I^\top \Delta\mW\|}{\| \mW_I \| \| \Delta\mW \|)})^2+\beta  (\frac{\|\Delta\mW^\top \mW_I\|}{\| \mW_I \| \| \Delta\mW \|})^2$ encourages orthogonality in both column and row spaces.
    \end{itemize}
        We tried different values of $\beta$ but found no significant change in the results, so we fix $\beta = 1$. 
    %\item 
    \item  \textbf{Post-hoc Orthogonalization:} We post-process $\Delta\mW$ before applying Equation~\ref{eq: lora}, projecting it onto the orthogonal complement of the top-$k$ singular vectors of $\mW_I$:
    \begin{itemize}[leftmargin=0.8cm]
\setlength\itemsep{0em}
        \item \texttt{OrthoMerge-col}: $\Delta\mW \leftarrow (\mI - \mU_k\mU_k^\top)\Delta\mW$ enforces column-space orthogonality based on the rank-$k$ SVD of $\mW_I$.

        \item \texttt{OrthoMerge-both}: $\Delta\mW \leftarrow (\mI - \mU_k\mU_k^\top)\Delta\mW(\mI - \mV_k\mV_k^\top)$ enforces orthogonality in both column and row spaces. We found that directly applying OrthoMerge-both leads to a drop in safety scores. To mitigate this, we further scale up the orthogonal complement with $    \Delta\mW \leftarrow \lambda(\mI - \mU_k\mU_k^\top)\Delta\mW(\mI - \mV_k\mV_k^\top) $ to compensates for the loss in safety. We experiment with different values of $\lambda$ in the range $\{ 1, 1.15, 1.75, 1.2, 1.25 \}$. We set $k=64$.
    \end{itemize}
% \end{itemize}
\vspace{-.2cm}
    \end{itemize}
For both approaches, we omit the row-space-only variant, as it showed no significant improvement in our experiments. 

Figure~\ref{fig:ortho_math_14b} shows the results on \texttt{AIME} and \texttt{GPQA} for the 14B models with $r=4$. For each method that yields multiple checkpoints (e.g., different epochs for vanilla and regularization finetuning, or different hyperparameters for \texttt{OrthoMerge}), we visualize the Pareto-frontier points. We observe that methods addressing both the row and column spaces (``both") tend to yield better results than those that only operate on the column space (``col"). Among them, \texttt{OrthoMerge-both} appears most promising, with more points concentrated in the upper-right corner and its best point strictly dominating vanilla LoRA.
 Therefore, we additionally evaluate \texttt{OrthoMerge-both} on the coding benchmarks in Figure~\ref{fig:ortho_code_14b}, where it slightly underperforms on \texttt{HumanEval+} but slightly outperforms vanilla on \texttt{MBPP+} in terms of the tradeoff.

Overall, we observe modest yet inconsistent gains from post-hoc orthogonalization. This highlights the potential of controlling the subspace geometry of LoRA updates and points to the need for more nuanced methods that yield consistent improvements.

\section{Complete Proofs for Section~\ref{sec:theory_linear}}
\label{apdx:lora_complete}

\subsection{Basic Lemmas}

\begin{lemma}[Population loss in closed form]
\label{lem:pop_loss_apdx}
Fix a task $\gT=(\gV,s,\sigma,\mTheta)$.
For any $\mW\in\sR^{d\times k}$,
\[
L(\mW,\gT)
=
s^2\bigl\|\mP_{\gV}(\mW-\mTheta)\bigr\|_F^2
+
\sigma^2\bigl\|\mP_{\gV^\perp}\mW\bigr\|_F^2.
\]
In particular, the unique minimizer is $\mW^\star=\mTheta$, and $L(\mW^\star,\gT)=0$.
\end{lemma}

\begin{proof}
Expand the prediction error:
\[
\vx^\top\mW-\vy
=
(\vz+\vxi)^\top\mW-\vz^\top\mTheta
=
\vz^\top(\mW-\mTheta)+\vxi^\top\mW.
\]
Since $\vz$ and $\vxi$ are independent mean-zero Gaussians, the cross-term vanishes, giving
\[
\E\|\vz^\top(\mW-\mTheta)+\vxi^\top\mW\|_2^2
=
\E\|\vz^\top(\mW-\mTheta)\|_2^2+\E\|\vxi^\top\mW\|_2^2.
\]
Using $\E[\vz\vz^\top]=s^2\mP_{\gV}$ and $\E[\vxi\vxi^\top]=\sigma^2\mP_{\gV^\perp}$,
\begin{align*}
\E\|\vz^\top(\mW-\mTheta)\|_2^2
&=\Tr\!\Big((\mW-\mTheta)^\top \E[\vz\vz^\top](\mW-\mTheta)\Big)
= s^2\|\mP_{\gV}(\mW-\mTheta)\|_F^2,\\
\E\|\vxi^\top\mW\|_2^2
&=\Tr\!\Big(\mW^\top \E[\vxi\vxi^\top]\mW\Big)
=\sigma^2\|\mP_{\gV^\perp}\mW\|_F^2.
\end{align*}
This proves the formula. Both terms are nonnegative and $\mP_{\gV}\mTheta=\mTheta$,
so the unique minimizer satisfies $\mP_{\gV}\mW=\mTheta$ and $\mP_{\gV^\perp}\mW=\mzero$, hence $\mW^\star=\mTheta$.
\end{proof}

\begin{lemma}[Eckart--Young \cite{eckart1936approximation}]
\label{lem:EY_apdx}
Let $\mM$ have singular values $\sigma_1(\mM)\ge \sigma_2(\mM)\ge\cdots\ge 0$.
Then
\[
\min_{\rank{\mX}\le p}\|\mM-\mX\|_F^2
=
\sum_{i>p}\sigma_i(\mM)^2,
\]
and a minimizer is the truncated SVD $\mM^{(p)}=\sum_{i=1}^p \sigma_i(\mM)\, \vu_i\vv_i^\top$.
\end{lemma}

\begin{lemma}[Rank splits across orthogonal subspaces]
\label{lem:rank_split_apdx}
Let $\mP$ be an orthogonal projector on $\sR^d$ and $\mP^\perp:=\mI-\mP$.
For any $\mS\in\sR^{d\times k}$,
the column spaces of $\mP\mS$ and $\mP^\perp\mS$ lie in orthogonal subspaces of $\sR^d$.
In particular,
\[
\rank{\mS}=\rank{\mP\mS}+\rank{\mP^\perp\mS}.
\]
\end{lemma}

\begin{proof}
$\mP\mS$ has columns in $\mathrm{range}(\mP)$ and $\mP^\perp\mS$ has columns in $\mathrm{range}(\mP^\perp)$,
which are orthogonal and intersect trivially. Hence
$\mathrm{col}(\mS)=\mathrm{col}(\mP\mS)\oplus \mathrm{col}(\mP^\perp\mS)$ and the rank splits.
\end{proof}

\subsection{Exact Characterization of LoRA Minimizers (Rank Allocation)}
\label{subsec:exact_characterization}

\begin{lemma}[LoRA finetuning minimizers admit an exact rank-allocation form]
\label{lem:lora_exact_apdx}
Assume $\gV_{\rm base}\perp \gV_{\rm ft}$, $\mP_{\gV_{\rm ft}}\mTheta_{\rm ft}=\mTheta_{\rm ft}$, and $\mP_{\gV_{\rm base}}\mTheta_{\rm base}=\mTheta_{\rm base}$.
Let $\mW_0=\mTheta_{\rm base}$ and consider
\[
\min_{\rank{\mS}\le r}\; L(\mW_0+\mS,\gT_{\rm ft}).
\]
For each $q\in\{0,1,\dots,r\}$ define
\[
F_r(q)
\;:=\;
s_{\rm ft}^2\sum_{i>q}\sigma_i(\mTheta_{\rm ft})^2
\;+\;
\sigma_{\rm ft}^2\sum_{i>r-q}\sigma_i(\mTheta_{\rm base})^2,
\qquad
q^*\in\arg\min_{0\le q\le r}F_r(q).
\]
Then every global minimizer $\mS^*$ satisfies
\[
\mP_{\gV_{\rm ft}}(\mW_0+\mS^*)=\mTheta_{\rm ft}^{(q^*)},
\qquad
\mP_{\gV_{\rm ft}^\perp}(\mW_0+\mS^*)=\mTheta_{\rm base}-\mTheta_{\rm base}^{(r-q^*)},
\]
and therefore
\[
\mW_{\rm lora}
=
\mW_0+\mS^*
=
\mTheta_{\rm ft}^{(q^*)}
+
\bigl(\mTheta_{\rm base}-\mTheta_{\rm base}^{(r-q^*)}\bigr).
\]
\end{lemma}

\begin{proof}
By Lemma~\ref{lem:pop_loss_apdx} applied to $\gT_{\rm ft}$ and $\gV_{\rm base}\perp \gV_{\rm ft}$ (so $\mP_{\gV_{\rm ft}}\mTheta_{\rm base}=\mzero$),
for any $\mS$,
\begin{align}
L(\mW_0+\mS,\gT_{\rm ft})
&=
s_{\rm ft}^2\bigl\|\mP_{\gV_{\rm ft}}(\mW_0+\mS-\mTheta_{\rm ft})\bigr\|_F^2
+
\sigma_{\rm ft}^2\bigl\|\mP_{\gV_{\rm ft}^\perp}(\mW_0+\mS)\bigr\|_F^2 \notag\\
&=
s_{\rm ft}^2\bigl\|\mP_{\gV_{\rm ft}}\mS-\mTheta_{\rm ft}\bigr\|_F^2
+
\sigma_{\rm ft}^2\bigl\|\mTheta_{\rm base}+\mP_{\gV_{\rm ft}^\perp}\mS\bigr\|_F^2.
\label{eq:ft_obj_decomp_apdx}
\end{align}
Let $\mX:=\mP_{\gV_{\rm ft}}\mS$ and $\mY:=\mP_{\gV_{\rm ft}^\perp}\mS$, so $\mS=\mX+\mY$.
By Lemma~\ref{lem:rank_split_apdx}, $\rank(\mS)=\rank(\mX)+\rank(\mY)$, hence the constraint $\rank(\mS)\le r$
is equivalent to $\rank(\mX)+\rank(\mY)\le r$.
Thus the problem becomes
\[
\min_{\rank(\mX)+\rank(\mY)\le r}
\Bigl[
s_{\rm ft}^2\|\mX-\mTheta_{\rm ft}\|_F^2
+\sigma_{\rm ft}^2\|\mTheta_{\rm base}+\mY\|_F^2
\Bigr].
\]
Fix $q\in\{0,\dots,r\}$ and impose $\rank(\mX)\le q$, $\rank(\mY)\le r-q$.
By Lemma~\ref{lem:EY_apdx},
\[
\min_{\rank(\mX)\le q}\|\mX-\mTheta_{\rm ft}\|_F^2=\sum_{i>q}\sigma_i(\mTheta_{\rm ft})^2,
\qquad
\min_{\rank(\mY)\le r-q}\|\mTheta_{\rm base}+\mY\|_F^2=\sum_{i>r-q}\sigma_i(\mTheta_{\rm base})^2.
\]
Therefore, the optimal value under split $q$ is exactly $F_r(q)$.
Minimizing over $q$ yields $q^*\in\arg\min F_r(q)$, and any global minimizer must achieve the Eckart--Young optima at $(q^*,r-q^*)$:
$\mX=\mTheta_{\rm ft}^{(q^*)}$ and $\mY=-\mTheta_{\rm base}^{(r-q^*)}$.
Substituting back gives the stated form of $\mW_{\rm lora}$.
\end{proof}

\subsection{Proof of Theorem~\ref{thm:full_forgets_main}}
\label{subsec:proof_full_forgets}

\begin{proof}
By Lemma~\ref{lem:pop_loss_apdx}, the unique minimizer of the finetuning population loss is $\mW_{\rm full}=\mTheta_{\rm ft}$ and
$L(\mW_{\rm full},\gT_{\rm ft})=0$.

Evaluate the base loss at $\mW_{\rm full}=\mTheta_{\rm ft}$:
\[
L(\mTheta_{\rm ft},\gT_{\rm base})
=
s_{\rm base}^2\|\mP_{\gV_{\rm base}}(\mTheta_{\rm ft}-\mTheta_{\rm base})\|_F^2
+
\sigma_{\rm base}^2\|\mP_{\gV_{\rm base}^\perp}\mTheta_{\rm ft}\|_F^2.
\]
Since $\gV_{\rm base}\perp \gV_{\rm ft}$ and $\mP_{\gV_{\rm ft}}\mTheta_{\rm ft}=\mTheta_{\rm ft}$, we have
$\mP_{\gV_{\rm base}}\mTheta_{\rm ft}=\mzero$ and $\mP_{\gV_{\rm base}^\perp}\mTheta_{\rm ft}=\mTheta_{\rm ft}$, and also
$\mP_{\gV_{\rm base}}\mTheta_{\rm base}=\mTheta_{\rm base}$. Hence
\[
L(\mTheta_{\rm ft},\gT_{\rm base})
=
s_{\rm base}^2\|\mTheta_{\rm base}\|_F^2
+
\sigma_{\rm base}^2\|\mTheta_{\rm ft}\|_F^2.
\]
Using normalization $s_{\rm base}^2\|\mTheta_{\rm base}\|_F^2=C$ gives
$L(\mW_{\rm full},\gT_{\rm base})\ge C$.
\end{proof}

\subsection{Proof of Theorem~\ref{thm:lora_ft_lb_main}}
\label{subsec:proof_ft_lb}

\begin{proof}
By Lemma~\ref{lem:pop_loss_apdx} applied to $\gT_{\rm ft}$, for any $\mW$,
\[
L(\mW,\gT_{\rm ft}) \;\ge\; s_{\rm ft}^2\|\mP_{\gV_{\rm ft}}(\mW-\mTheta_{\rm ft})\|_F^2.
\]
Under $\gV_{\rm base}\perp \gV_{\rm ft}$ and $\mW_0=\mTheta_{\rm base}$, we have $\mP_{\gV_{\rm ft}}\mW_0=\mzero$.
Any LoRA model satisfies $\mW=\mW_0+\mS$ with $\rank(\mS)\le r$, hence $\rank(\mP_{\gV_{\rm ft}}\mW)=\rank(\mP_{\gV_{\rm ft}}\mS)\le r$.
Therefore
\[
L(\mW_{\rm lora},\gT_{\rm ft})
\;\ge\;
s_{\rm ft}^2\min_{\rank(\mX)\le r}\|\mX-\mTheta_{\rm ft}\|_F^2
=
s_{\rm ft}^2\sum_{i>r}\sigma_i(\mTheta_{\rm ft})^2,
\]
where the equality is Lemma~\ref{lem:EY_apdx}.
Next,
\[
\sum_{i>r}\sigma_i(\mTheta_{\rm ft})^2
=
\|\mTheta_{\rm ft}\|_F^2-\sum_{i\le r}\sigma_i(\mTheta_{\rm ft})^2
\ge
\|\mTheta_{\rm ft}\|_F^2-r\|\mTheta_{\rm ft}\|_2^2
=
\|\mTheta_{\rm ft}\|_F^2\Bigl(1-\frac{r}{\intdim(\mTheta_{\rm ft})}\Bigr).
\]
Truncating at zero and using $s_{\rm ft}^2\|\mTheta_{\rm ft}\|_F^2=C$ yields
\[
L(\mW_{\rm lora},\gT_{\rm ft})
\;\ge\;
C\Bigl(1-\frac{r}{\intdim(\mTheta_{\rm ft})}\Bigr)_+.
\]
\end{proof}

\subsection{Proof of Theorem~\ref{thm:lora_tradeoff_main}}
\label{subsec:proof_tradeoff}

\begin{proof}
Let $r_{\rm ft}:=\rank(\mTheta_{\rm ft})$ and assume $r\ge r_{\rm ft}$.
We first show that Assumption~\ref{as:spectral_sep_main} implies the rank-allocation optimizer in Lemma~\ref{lem:lora_exact_apdx}
satisfies $q^*=r_{\rm ft}$.
Fix $q<r_{\rm ft}$ and let $t:=r_{\rm ft}-q\ge 1$. Write $\Delta=r-r_{\rm ft}$, so $r-q=\Delta+t$.
Using the definition of $F_r(\cdot)$ from Lemma~\ref{lem:lora_exact_apdx},
\begin{align*}
F_r(q)-F_r(r_{\rm ft})
&=
s_{\rm ft}^2\sum_{i=q+1}^{r_{\rm ft}}\sigma_i(\mTheta_{\rm ft})^2
-
\sigma_{\rm ft}^2\sum_{i=\Delta+1}^{\Delta+t}\sigma_i(\mTheta_{\rm base})^2\\
&\ge
t\Bigl(s_{\rm ft}^2\sigma_{r_{\rm ft}}(\mTheta_{\rm ft})^2-\sigma_{\rm ft}^2\sigma_{1}(\mTheta_{\rm base})^2\Bigr)
\;>\;0,
\end{align*}
where the final strict inequality is exactly Assumption~\ref{as:spectral_sep_main} after squaring.
Hence no $q<r_{\rm ft}$ can be optimal, and since $r\ge r_{\rm ft}$, we obtain $q^*=r_{\rm ft}$.

With $q^*=r_{\rm ft}$, Lemma~\ref{lem:lora_exact_apdx} gives
\[
\mW_{\rm lora}
=
\mTheta_{\rm ft}
+
\bigl(\mTheta_{\rm base}-\mTheta_{\rm base}^{(\Delta)}\bigr),
\qquad
\Delta=r-r_{\rm ft}.
\]

\paragraph{Finetuning loss.}
By Lemma~\ref{lem:pop_loss_apdx} for $\gT_{\rm ft}$ and $\gV_{\rm base}\perp \gV_{\rm ft}$,
\[
L(\mW_{\rm lora},\gT_{\rm ft})
=
\sigma_{\rm ft}^2\|\mTheta_{\rm base}-\mTheta_{\rm base}^{(\Delta)}\|_F^2
\le
\sigma_{\rm ft}^2\|\mTheta_{\rm base}\|_F^2
=
\sigma_{\rm ft}^2\cdot \frac{C}{s_{\rm base}^2}
=
o(1),
\]
using normalization and Assumption~\ref{as:noise_sep_main}.

\paragraph{Base loss.}
By Lemma~\ref{lem:pop_loss_apdx} for $\gT_{\rm base}$ and $\mP_{\gV_{\rm base}}\mTheta_{\rm ft}=\mzero$,
\[
L(\mW_{\rm lora},\gT_{\rm base})
=
s_{\rm base}^2\|\mTheta_{\rm base}^{(\Delta)}\|_F^2
+
\sigma_{\rm base}^2\|\mTheta_{\rm ft}\|_F^2
=
s_{\rm base}^2\|\mTheta_{\rm base}^{(\Delta)}\|_F^2
+
o(1),
\]
again by normalization and Assumption~\ref{as:noise_sep_main}.
Finally,
\[
\|\mTheta_{\rm base}^{(\Delta)}\|_F^2
=
\sum_{i\le \Delta}\sigma_i(\mTheta_{\rm base})^2
\le
\Delta\,\sigma_1(\mTheta_{\rm base})^2
=
\Delta\,\|\mTheta_{\rm base}\|_2^2,
\]
so
\[
L(\mW_{\rm lora},\gT_{\rm base})
\le
s_{\rm base}^2\Delta\|\mTheta_{\rm base}\|_2^2
+
o(1)
=
C\cdot\frac{\Delta}{\intdim(\mTheta_{\rm base})}
+
o(1),
\]
since $C=s_{\rm base}^2\|\mTheta_{\rm base}\|_F^2$ and $\intdim(\mTheta_{\rm base})=\|\mTheta_{\rm base}\|_F^2/\|\mTheta_{\rm base}\|_2^2$.
\end{proof}

\subsection{Lower Bounds for the Base Loss}
\label{subsec:base_lb}

\begin{corollary}[A rank-based lower bound on overshoot cost]
\label{cor:base_lb_rank}
Assume the conditions of Theorem~\ref{thm:lora_tradeoff_main} and write $\Delta=r-r_{\rm ft}$.
Let $R_{\rm base}:=\rank(\mTheta_{\rm base})$.
Then
\[
L(\mW_{\rm lora},\gT_{\rm base})
\;\ge\;
C\cdot\frac{\Delta}{R_{\rm base}}
+o(1).
\]
\end{corollary}

\begin{proof}
Under Theorem~\ref{thm:lora_tradeoff_main}, we have
$L(\mW_{\rm lora},\gT_{\rm base})=s_{\rm base}^2\sum_{i\le \Delta}\sigma_i(\mTheta_{\rm base})^2+o(1)$.
For any nonincreasing nonnegative sequence $(a_1,\dots,a_{R_{\rm base}})$,
$\sum_{i\le \Delta}a_i\ge \frac{\Delta}{R_{\rm base}}\sum_{i\le R_{\rm base}}a_i$.
Applying this to $a_i=\sigma_i(\mTheta_{\rm base})^2$ yields
$\sum_{i\le \Delta}\sigma_i^2\ge \frac{\Delta}{R_{\rm base}}\|\mTheta_{\rm base}\|_F^2$.
Multiplying by $s_{\rm base}^2$ and using $s_{\rm base}^2\|\mTheta_{\rm base}\|_F^2=C$ gives the result.
\end{proof}

\begin{assumption}[Flat spectrum for the base task (optional)]
\label{as:flat_base}
There exists $\kappa\ge 1$ such that
\[
\sigma_i(\mTheta_{\rm base})^2 \;\ge\; \frac{1}{\kappa}\,\sigma_1(\mTheta_{\rm base})^2
\qquad\text{for all }1\le i\le \lfloor \intdim(\mTheta_{\rm base})\rfloor.
\]
\end{assumption}

\begin{corollary}[With flat spectrum: a matching $\intdim$-based lower bound]
\label{cor:base_lb_intdim}
Assume the conditions of Theorem~\ref{thm:lora_tradeoff_main} and Assumption~\ref{as:flat_base}.
Let $\Delta=r-r_{\rm ft}$ and $d_{\rm base}:=\intdim(\mTheta_{\rm base})$.
Then
\[
L(\mW_{\rm lora},\gT_{\rm base})
\;\ge\;
\frac{C}{\kappa}\cdot \min\Bigl\{\frac{\Delta}{d_{\rm base}},\,1\Bigr\}
+o(1).
\]
In particular, for $\Delta\le d_{\rm base}$,
$L(\mW_{\rm lora},\gT_{\rm base})\ge \frac{C}{\kappa}\cdot \frac{\Delta}{d_{\rm base}}+o(1)$,
matching the scaling in Theorem~\ref{thm:lora_tradeoff_main} up to the factor $\kappa$.
\end{corollary}

\begin{proof}
Under Theorem~\ref{thm:lora_tradeoff_main},
$L(\mW_{\rm lora},\gT_{\rm base})=s_{\rm base}^2\sum_{i\le \Delta}\sigma_i(\mTheta_{\rm base})^2+o(1)$.
If $\Delta\le \lfloor d_{\rm base}\rfloor$, Assumption~\ref{as:flat_base} gives
$\sum_{i\le \Delta}\sigma_i^2 \ge \Delta\cdot \sigma_1^2/\kappa$.
Using $\sigma_1^2=\|\mTheta_{\rm base}\|_F^2/d_{\rm base}$ yields
$\sum_{i\le \Delta}\sigma_i^2 \ge (\Delta/(\kappa d_{\rm base}))\|\mTheta_{\rm base}\|_F^2$.
Multiplying by $s_{\rm base}^2$ and using $s_{\rm base}^2\|\mTheta_{\rm base}\|_F^2=C$ gives the desired bound.
If $\Delta>\lfloor d_{\rm base}\rfloor$, then $\sum_{i\le \Delta}\sigma_i^2\ge \sum_{i\le \lfloor d_{\rm base}\rfloor}\sigma_i^2
\ge \lfloor d_{\rm base}\rfloor \sigma_1^2/\kappa \gtrsim \|\mTheta_{\rm base}\|_F^2/\kappa$,
which yields the saturation at $\gtrsim C/\kappa$ (absorbing flooring into constants and $o(1)$).
\end{proof}

\subsection{Discussion on Gradient Descent}\label{apdx:gd}

Recall the population loss on a task $\gT=(\gV,s,\sigma,\mTheta)$:
\[
L(\mW,\gT)=\E\bigl\|\vx^\top \mW-\vy\bigr\|_2^2,
\qquad \vx=\vz+\vxi,\ \vz\sim \mathcal{N}(0,s^2 \mP_{\gV}),\ \vxi\sim \mathcal{N}(0,\sigma^2 \mP_{\gV^\perp}),
\qquad \vy=\vz^\top \mTheta.
\]
Let $\mSigma_x := \E[\vx\vx^\top]= s^2 \mP_{\gV}+\sigma^2 \mP_{\gV^\perp}$ and $\mSigma_{xy}:=\E[\vx\vy^\top]=s^2\mP_{\gV}\mTheta$.
Then
\begin{equation}
\label{eq:population_quadratic_form}
L(\mW,\gT)= \Tr(\mW^\top \mSigma_x \mW)-2\Tr(\mW^\top \mSigma_{xy})+\Tr(\E[\vy\vy^\top]).
\end{equation}
By Lemma~\ref{lem:pop_loss_apdx}, the minimizer of $L(\mW,\gT)$ is unique and is given by $\mW^\star=\mTheta$.

\paragraph{Full-model finetuning.}
Consider the finetuning task $\gT_{\rm ft}=(\gV_{\rm ft},s_{\rm ft},\sigma_{\rm ft},\mTheta_{\rm ft})$ with $s_{\rm ft}>0$ and $\sigma_{\rm ft}>0$.
We run gradient descent on $\mW$:
\[
\mW_{t+1}
=\mW_t-\eta\,\nabla_{\mW}L(\mW_t,\gT_{\rm ft})
=\mW_t-2\eta\bigl(\mSigma_x^{\rm ft}\mW_t-\mSigma_{xy}^{\rm ft}\bigr).
\]
Let $\lambda_{\max}:=\lambda_{\max}(\mSigma_x^{\rm ft})=\max\{s_{\rm ft}^2,\sigma_{\rm ft}^2\}$.
If $0<\eta<1/\lambda_{\max}$, then gradient descent converges linearly to the unique minimizer $\mW_{\rm full}=\mTheta_{\rm ft}$:
\[
\mW_t-\mW_{\rm full}
=
(\mI-2\eta\,\mSigma_x^{\rm ft})^t(\mW_0-\mW_{\rm full}),
\]
and hence $\mW_t\to\mW_{\rm full}$ as $t\to\infty$.

\paragraph{LoRA finetuning.} In the population setting, the finetuning loss is a quadratic function of $\mW$:
\[
L(\mW,\gT_{\rm ft})
=\Tr(\mW^\top \mSigma_x^{\rm ft}\mW)-2\Tr(\mW^\top \mSigma_{xy}^{\rm ft})+\Tr(\E[\vy\vy^\top]),
\]
where $\mSigma_x^{\rm ft}\succ 0$ since $s_{\rm ft},\sigma_{\rm ft}>0$. Completing the square gives
\[
L(\mW,\gT_{\rm ft})
=
\bigl\|(\mSigma_x^{\rm ft})^{1/2}\mW-(\mSigma_x^{\rm ft})^{-1/2}\mSigma_{xy}^{\rm ft}\bigr\|_F^2
+\text{const}.
\]
With the LoRA parameterization $\mW=\mW_0+\mB\mA$, define
\[
\mM:=(\mSigma_x^{\rm ft})^{-1/2}\mSigma_{xy}^{\rm ft}-(\mSigma_x^{\rm ft})^{1/2}\mW_0.
\]
Then minimizing $L(\mW_0+\mB\mA,\gT_{\rm ft})$ over $(\mB,\mA)$ is equivalent (up to an additive constant and an invertible linear change of variables in $\mB$)
to the classical matrix factorization / shallow linear network objective
\[
\min_{\mB\in\R^{d\times r},\,\mA\in\R^{r\times k}} \ \|(\mSigma_x^{\rm ft})^{1/2}\mB\mA-\mM\|_F^2.
\]
Classical results in shallow linear networks / matrix factorization imply that this objective has a benign landscape.
To see this, let $\mC\coloneqq(\mSigma_x^{\rm ft})^{1/2}\succ 0$ and reparameterize $\mU\coloneqq \mC\mB$.
Since $\mC$ is invertible, the map $(\mB,\mA)\mapsto(\mU,\mA)$ is a bijection and
\[
\min_{\mB,\mA}\ \|\mC\mB\mA-\mM\|_F^2
\quad\equiv\quad
\min_{\mU,\mA}\ \|\mU\mA-\mM\|_F^2,
\]
which is the classical matrix factorization / shallow linear network objective.
Moreover, invertible linear changes of variables preserve critical points and the presence of negative curvature
(i.e., $(\mB,\mA)$ is critical iff $(\mU,\mA)$ is critical, and the Hessian has a negative eigenvalue at one iff it
does at the other). For this objective, every local minimum is a global minimum, and every critical point that is not globally optimal
is a saddle point \citep{baldi1989neural,kawaguchi2016deep,valavi2020revisiting,zhu2021global}. Consequently, to argue global optimality of gradient descent, it suffices to show that it converges to a local minimum rather than a saddle. Under the strict-saddle geometry, gradient descent with random initialization
(almost surely) does not converge to strict saddles and instead converges to a local minimizer
\citep{lee2016gradient,panageas2016gradient}. Together with the absence of spurious local minima, this implies that gradient descent on the LoRA parameters converges to a global minimizer under mild conditions.

\end{document}